\documentclass[10pt,journal,compsoc]{IEEEtran}

\ifCLASSOPTIONcompsoc
  \usepackage[nocompress]{cite}
\else
  \usepackage{cite}
\fi
\ifCLASSINFOpdf
\else
\fi

\usepackage{wrapfig}
\usepackage{epsfig}
\usepackage{setspace}
\usepackage{varwidth}
\usepackage{microtype}
\usepackage{graphicx}
\usepackage{subfigure}
\usepackage{changepage}
\usepackage{multirow}
\usepackage{tabularx}
\usepackage{makecell}
\usepackage{tikz}
\usepackage{pifont}
\usepackage{color}
\usepackage{enumerate}
\usepackage{enumitem}
\usepackage{caption}
\usepackage{diagbox}
\usepackage{parskip}
\usepackage{soul}
\usepackage{pifont}
\usepackage{booktabs}
\usepackage[numbers,sort&compress]{natbib}
\usepackage{caption}
\definecolor{tred}{RGB}{128,0,32}
\definecolor{tgreen}{RGB}{34,139,34}

\usepackage{booktabs} 

\newcommand{\blackcircle}[1]{%
  \tikz[baseline=(char.base)]{
    \node[shape=circle,fill=black,inner sep=0.2pt] (char) {\textcolor{white}{#1}};}}
    
\usepackage{amsmath,amsfonts,bm}

\def\eqref#1{equation~\ref{#1}}

\def\1{\bm{1}}

\def\ve{{\bm{e}}}

\def\vp{{\bm{p}}}

\def\vt{{\bm{t}}}

\def\mZ{{\bm{Z}}}

\DeclareMathAlphabet{\mathsfit}{\encodingdefault}{\sfdefault}{m}{sl}
\SetMathAlphabet{\mathsfit}{bold}{\encodingdefault}{\sfdefault}{bx}{n}
\newcommand{\tens}[1]{\bm{\mathsfit{#1}}}
\def\tA{{\tens{A}}}

\def\tG{{\tens{G}}}

\def\tI{{\tens{I}}}

\def\tM{{\tens{M}}}

\def\tS{{\tens{S}}}
\def\tT{{\tens{T}}}

\def\sY{{\mathbb{Y}}}

\newcommand{\R}{\mathbb{R}}

\usepackage[utf8]{inputenc} 
\usepackage[T1]{fontenc}    
\usepackage[pagebackref=false,breaklinks=true,colorlinks,bookmarks=false]{hyperref}
\usepackage{url}            
\usepackage{booktabs}       
\usepackage{amsfonts}       
\usepackage{nicefrac}       
\usepackage{microtype}      
\usepackage{xcolor}         
\usepackage{amsmath}
\usepackage{amssymb}
\usepackage{mathtools}
\usepackage{amsthm}

\begin{document}
%
\title{Decouple before Align: Visual Disentanglement Enhances Prompt Tuning}
%
%
%
%

\author{Fei Zhang,
        Tianfei Zhou,
        Jiangchao Yao,
        Ya Zhang, 	
        Ivor W. Tsang, ~\IEEEmembership{Fellow, ~IEEE},
        Yanfeng Wang
\IEEEcompsocitemizethanks{\IEEEcompsocthanksitem F. Zhang, J. Yao are with Cooperative Medianet Innovation Center, Shanghai Jiao Tong University, Shanghai 200240, China.  Y. Zhang and Y. Wang are with School of Artificial Intelligence, Shanghai Jiao Tong University, Shanghai 200230, China. F. Zhang is also with Shanghai Innovation Institute. J. Yao, Y. Zhang and Y. Wang are also with Shanghai Artificial Intelligence Laboratory. The corresponding authors are Jiangchao Yao and Ya Zhang. \protect


E-mail: \{ferenas, Sunarker,ya\_zhang, wangyanfeng622\}@sjtu.edu.cn.
\IEEEcompsocthanksitem T. Zhou is with Beijing Institute of Technology, China. 
\IEEEcompsocthanksitem Ivor W. Tsang is with the A*STAR Centre for Frontier AI Research, Singapore.
}

}

%
%

\markboth{Journal of \LaTeX\ Class Files,~Vol.~14, No.~8, October~2024}%
{Shell \MakeLowercase{\textit{et al.}}: Decouple before Align: Visual Disentanglement Enhances Prompt Tuning}
%



\IEEEtitleabstractindextext{%
\begin{abstract}
\emph{Prompt tuning} (PT), as an emerging resource-efficient fine-tuning paradigm, has showcased remarkable effectiveness in improving the task-specific transferability of \emph{vision-language models}. This paper delves into a previously overlooked \emph{information asymmetry} issue in PT, where the visual modality mostly conveys more context than the object-oriented textual modality. Correspondingly, coarsely aligning these two modalities could result in the \emph{biased attention}, driving the model to merely focus on the context area. To address this, we propose DAPT, an effective PT framework based on an intuitive \emph{decouple-before-align} concept. First, we propose to explicitly decouple the visual modality into the foreground and background representation via exploiting coarse-and-fine visual segmenting cues, and then both of these decoupled patterns are aligned with the original foreground texts and the hand-crafted background classes, thereby symmetrically strengthening the modal alignment. To further enhance the visual concentration, we propose a visual pull-push regularization tailored for the foreground-background patterns, directing the original visual representation towards unbiased attention on the \emph{region-of-interest} object.  We demonstrate the power of architecture-free DAPT through \emph{few-shot learning}, \emph{base-to-novel generalization}, and \emph{data-efficient learning}, all of which yield superior performance across prevailing benchmarks. Our code will be released at \url{https://github.com/Ferenas/DAPT}.
\end{abstract}

}

\maketitle

\IEEEdisplaynontitleabstractindextext

%
\IEEEpeerreviewmaketitle

\ifCLASSOPTIONcompsoc
\IEEEraisesectionheading{\section{Introduction}\label{sec:introduction}}
\else
\section{Introduction}
\label{sec:introduction}
\fi

%
%
%
%
\begin{quote}
\vspace{-0.5mm}
``\ul{\emph{A picture is worth a thousand words}}.'' \\ 
\raggedleft
{--Brisbane Arthur}
\vspace{-1mm}
\end{quote}

\IEEEPARstart{T}{he} emerging \emph{vision-language foundation models} (VLMs), such as CLIP~\cite{clip} and BLIP~\cite{li2022blip}, have made a transformative impact on the field of artificial intelligence due to their powerful ability to generalize across various concepts. These CLIP-based models, through the use of a simple crafted prompt for the query class (e.g., ``\emph{a photo of a} [\texttt{CLASS NAME}]''), showcase impressive zero-shot recognition capabilities for numerous downstream tasks ~\cite{gu2021open,rao2022denseclip}.

Regardless of such powerful generalization, there has been significant interest from both academia and industry in tailoring these CLIP-based VLMs towards more promising task-specific performance through  \emph{prompt tuning} (PT). PT is a resource-efficient fine-tuning paradigm originally designed for \emph{large language models} (LLMs)~\cite{bert,gpt3}, and recent advances~\cite{coop, cocoop, vpt, prograd, maple} have extended the utility of such a tuning mechanism on CLIP by incorporating a few learnable prompt tokens to the textual/visual input embeddings. As task-specific-optimized prompts tend to overfit the tuning domain accompanied by losing the original generalization capabilities, the majority of these works have mainly focused on designing effective prompt regularizations to learn a well-balanced feature representation in both task-specific learning and novel-domain generalization.

\begin{figure}[t]
    \includegraphics[width=1.0\linewidth]{ 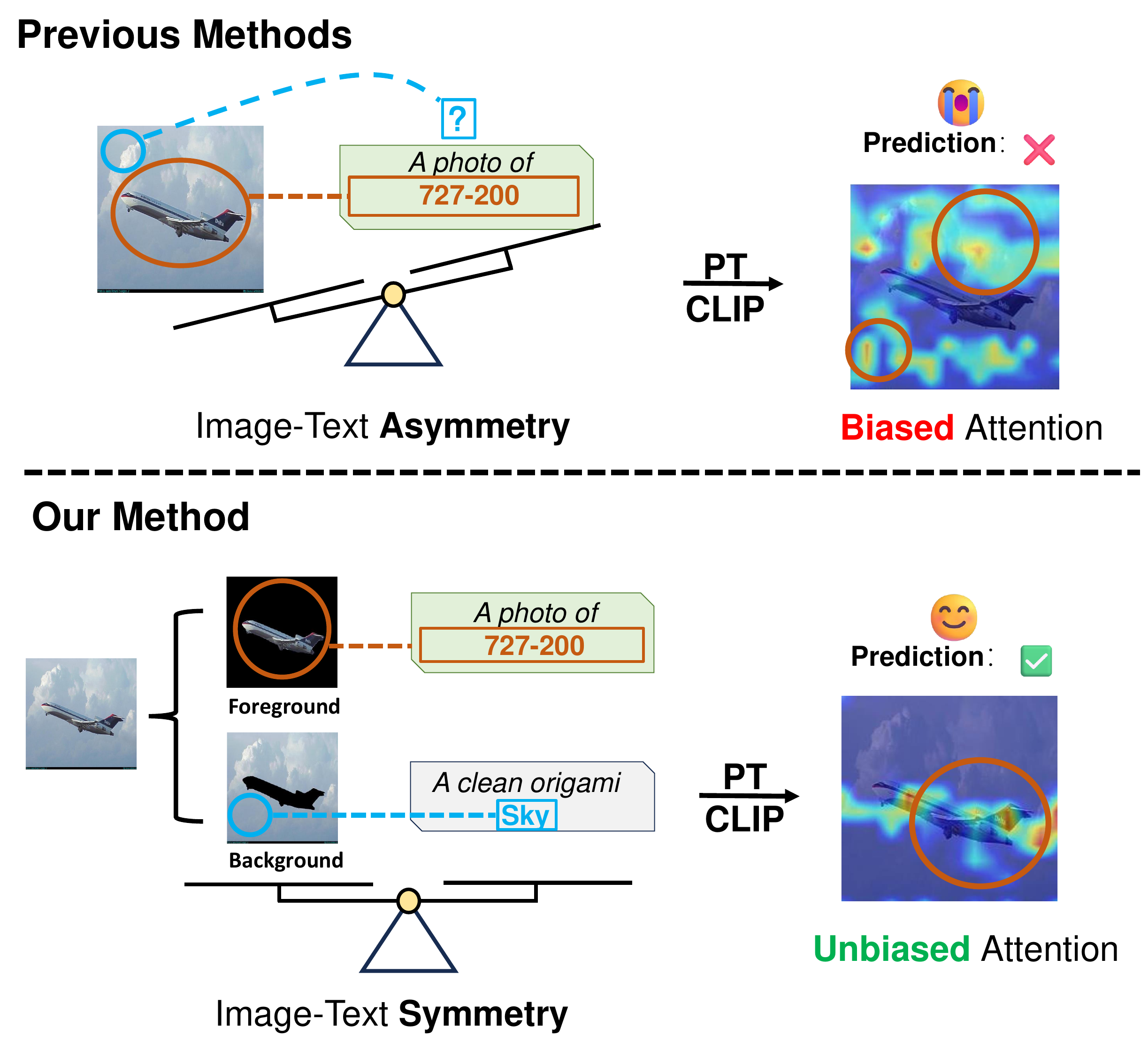}
    \vspace{-5mm}
    \caption{Illustration of our motivation. Compared to previous methods, which lead to \emph{biased attention} towards non-interest regions by overlooking the \emph{information asymmetric} within the misclassified samples, our method achieves a symmetrical image-text alignment by decoupling the visual and textual pattern, which directs CLIP focus on the ROI to perform accurate recognition.}
    \label{fig_teaser}
\end{figure}

Despite persistent advancements, these methods overlook a fundamental discrepancy between PT \textit{w.r.t.} VLMs and PT \textit{w.r.t.} LLMs—the issue of \emph{information asymmetry} in the image-text alignment. Unlike LLMs, where manipulating the textual modality is the sole option for semantic expression, VLMs possess an additional visual modality that naturally contains rich semantics: \ul{an image, mostly containing non-interest objects, could convey far more information, e.g., background context, than a text simply describing the visual interest.} Consequently, attempting to align these two information-asymmetric modalities can easily result in the \emph{biased attention}. As illustrated in Figure~\ref{fig_teaser}, the model with coarse image-text alignment tends to focus merely on the relevant contexts while neglecting the \emph{region-of-interest} (ROI) object. To address this, this paper aims to explicitly bridge the cross-modal information gap by symmetrizing the semantic patterns in both the visual and textual modalities, guiding the model towards more accurate recognition.

Accordingly, such modality asymmetry inherently stems from an overload of visual information, driving us to consider whether \emph{the visual representation could be explicitly decoupled} to direct a symmetric image-text alignment. \cite{shtedritski2023does, yang2023set} have revealed that the VLMs could demonstrate an emergent fine-grained recognition ability by highlighting the ROI in an image through a set of \emph{visual cues}, e,g., a circle and object-wise mask.  Motivated by this, we aim to explore and exploit such a straightforward concept to shift the attention of CLIP towards object-oriented textual prompts, setting the stage for achieving modal symmetry by establishing bijective and redundancy-free image-text correlation for PT.

To this end, we propose the \emph{visual disentanglement} that partitions the image input into the foreground and background part by a semantic mask, where the foreground serves as the highlighted region corresponding to the text prompt. Specifically, we explore two types of semantic masks depending on their generation sources, i.e., Grad-CAM~\cite{gradcam} and SEEM~\cite{seem}. The former is a model-self-driven coarse attention mask, while the latter is an external segmentation model crafting fine-grained masks. These approaches furnish dual coarse-and-fine strategies for visual decoupling. Accordingly, we propose effective visual and textual regularizations to perform symmetrical modal alignment for PT. Firstly, we propose the \emph{foreground-text alignment} that tailors the attention of CLIP to the textual object. To leverage the context knowledge of the background, we further introduce a certain number of background classes to perform the \emph{background-text alignment}, explicitly enhancing model generalization. Then, to explicitly alter the model's attention, we propose the \emph{visual triplets mining} that, through a pull-push triplet loss, pulls the prompted feature of the original image close to the foreground while pushing it away from the background. Based on these regularized items, we propose DAPT, a \emph{decouple-before-align} PT framework that strengthens the recognition capability of CLIP against in- and out-of domains. Overall, we make the following contributions:
\begin{enumerate}[leftmargin=*,label=\textbf{\textbullet}]
\item {We propose the \emph{visual disentanglement} that exploits the \emph{visual cues} of different levels to highlight the text-oriented object in the visual modality. This explicit accentuation is encouraged to alter the attention of CLIP towards an accurately-recognized pattern, addressing the \emph{biased attention} led from the asymmetrical image-text alignment.}
\item We propose DAPT, a simple yet effective prompting architecture that performs visual pull-push regularization, and bijective image-text \emph{alignment} with the \emph{decoupled} visual and textual patterns, injecting symmetrical modality information for CLIP to improve the effectiveness of PT. 
\item Extensive results on quantitative benchmarks demonstrate the effectiveness of DAPT, yielding new \emph{state-of-the-art} (SOTA) performance on both task-specific learning and base-to-novel generalization. Particularly, DAPT could, with saving about \textbf{50}\% training data, achieve comparable performance against other methods, which further shows the superiority of DAPT in data-efficient learning.
\end{enumerate}

\section{Related Work}
\subsection{Prompt Tuning for VLMs}
PT, originally fit for LLMs~\cite{bert, gpt3} to achieve quick domain adaptation, has been circumstantially investigated for the CLIP~\cite{clip}-based VLMs to benefit the downstream tasks with merely a few learnable trainable parameters. CoOp~\cite{coop} and CoCoOp~\cite{cocoop} pave the way for PT in CLIP, by optimizing a set of learnable token embeddings at the textual input. Based on this, a series of advances~\cite{prograd,chen2022plot,yao2023visual,guo2023texts} have been proposed to explore efficient PT frameworks for CLIP on the text-oriented pipeline. ~\cite{prograd} proposed a gradient-based optimization regularization to relieve the forgetting issue in PT. Another line of works~\cite{vpt} have shed light on the image-oriented prompt optimization, where the learnable prompts are concatenated to the visual embebddings. To fully exploit the multi-modal knowledge for PT, recent works~\cite{maple,selfregulating,promptkd,cao2024domain,cao2025generalized} have explored multi-modal prompts on both the visual and textual side of CLIP, showing robust and superior transferring ability. \cite{maple} proposed to learn hierarchical prompts jointly at the vision and language branches of CLIP to further improve the adaptation performance. To achieve balanced performance across base-to-novel game, \cite{selfregulating,promptkd} have turned to regulating the prompted representations with the frozen CLIP in case of overfitting. Our work focusing on the \emph{information asymmetry} issue is orthogonal to these explorations. 

\begin{figure*}[t]
    \centering
    \includegraphics[width=1.0\textwidth]{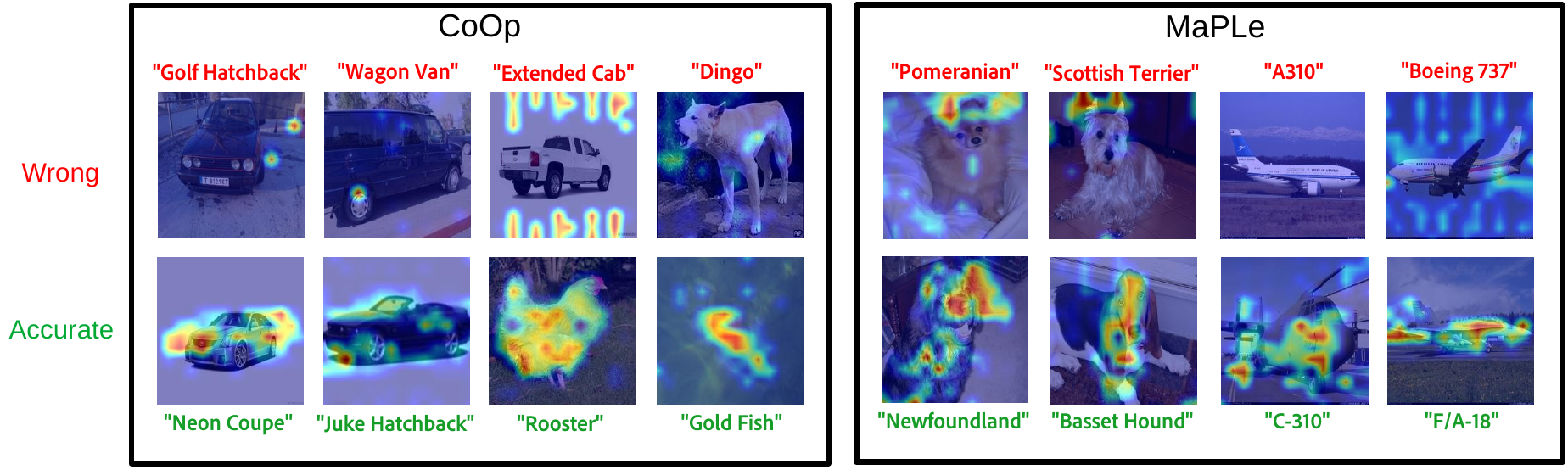}
    \vspace{-6mm}
    \caption{Illustrative attention of \textcolor{green}{accurately}/\textcolor{red}{wrongly}-classified samples from CoOp~\cite{coop} and MaPLe~\cite{maple}.  Intuitively, the
salient foreground attention reflects the accurate pattern learned from model, but the \emph{biased attention}, including no or few ROI activation towards the misclassified samples, reveals the inferior fine-grained recognition from simple image-text alignment.}
    \label{fig_illustration}
\end{figure*}

\subsection{Explicit Visual Cues for Prompting}
Different from parameterized prompts in PT, \emph{visual cues}, as special visual hints directly on the images with the forms of, e.g., a circle, bounding box, or a point, could also efficiently prompt vision-based foundation models in an intuitive manner~\cite{kirillov2023segment, seem, zhang2021complementary, zhang2023uncovering, ma2023attrseg, ma2023diffusionseg, ma2023open, chen2024probabilistic,zhang2025context, zhang2025g4seg}. Inspired by this, recent works have adopted this mechanism in tuning VLMs by developing the visual marks, e.g., red circle~\cite{shtedritski2023does}, a highlighted region~\cite{yang2024fine}, or a fine-grained object mask~\cite{yang2023set}. Particularly, FGVP~\cite{yang2024fine} adopts a generated mask contour by powerful off-the-shelf segmentation tool to implement Gaussian Blurring for the background, improving the dense perception of VLM towards the query foreground. To seamlessly incorporate these visual prompts, ~\cite{sun2024alpha} proposed to introduce an extra alpha channel for the input images, which suggests the attentive
regions by using segmentation masks. Except for these intuitive prompts, ~\cite{cai2024vip} proposed to introduce various flexible prompts, e.g., red arrows, for better human interaction. Remarkably, taking advantage of these \emph{visual cues} has been demonstrated to invoke the potential of VLMs in fine-grained and localized recognition capability. 
Motivated by this, this paper leverages this explicit mechanism for visual decoupling to explicitly improve the vision-language alignment.

\section{Preliminaries}\label{sec_preliminiary}

\noindent\textbf{Zero-shot inference on CLIP.} Formally, CLIP is comprised of two parallel encoders, denoted as $\mathcal{F}_{\mathrm{I}}: \R^{b \times 3 \times h \times w} \rightarrow \R^{b \times d}$ and $\mathcal{F}_{\mathrm{T}}: \R^{b \times l \times c_{\mathrm{T}}} \rightarrow \R^{b \times d}$, that maps $b$ images $ \{\tI^{i} \in \R^{3 \times h \times w} \}_{i=1}^{b} $ and texts $\{\tT^{i} \in \R^{l \times c_{\mathrm{T}}} \}_{i=1}^{b}$ into the visual and textual latent features, denoted as $\mZ_{\mathrm{I}} = \mathcal{F}_{\mathrm{I}}(\tI) \in \R^{1 \times d}$ and $\mZ_{\mathrm{T}} = \mathcal{F}_{\mathrm{T}}(\tT) \in \R^{1 \times d}$, respectively. Here, $h$ and $w$ denote the height and width of an image, $l$ and $c_\mathrm{T}$ denote the length and the tokenized dimension of a text embedding, and $d$ denotes the feature dimension. Note that an image, before feeding to $\mathcal{F}_{\mathrm{I}}$, is divided into $n$ patches to sequentially generate the image embedding $\tI = \{\texttt{CLS}, \ve_1,..., \ve_n\}$, where $\texttt{CLS}$ is an extra token for global visual representation, $\ve \in \R^{1 \times c_{\mathrm{I}}}$ denotes the patch embedding, and $c_\mathrm{I}$ denotes the dimension of image embedding. Similarly, the text embedding could be formulated as $\tT = \{\vt_1,..., \vt_l\} \in \R^{l \times c_\mathrm{T}}$, where $\vt \in \R^{1 \times c_\mathrm{T}}$ refers to the word embedding. After being pre-trained, CLIP could perform zero-shot inference on any downstream classification tasks based on an intuitive image-text matching problem. Specifically, suppose a $k$-classification problem and let $\sY = \{1,..,k\}$ denote the label space. For each class, we define the prompt template as ``\emph{a photo of a} [\texttt{CLASS NAME}]'' to form all the labels into $k$ textual descriptions. Then, the prediction problem could be defined as 
\begin{equation}
p(y|\mZ_{\mathrm{I}}, \mZ_{\mathrm{T}}) = \frac {\exp(\texttt{sim}(\mZ_{\mathrm{I}},  \mZ_{\mathrm{T}}^{y}))} {\sum\nolimits_{y=1}^{k} \exp(\texttt{sim}(\mZ_{\mathrm{I}}, \mZ_{\mathrm{T}}^{y}))},
\end{equation}
where $\texttt{sim}(\cdot,\cdot)$ denotes the cosine similarity score, and $\{\mZ_{\mathrm{T}}^{y}\}_{y=1}^k$ denotes all class-wise textual features. The predicted result is equivalent to the maximum class score.

\noindent\textbf{Prompt Tuning on CLIP.} PT, while keeping $\mathcal{F}_{\mathrm{I}}$ and $\mathcal{F}_{\mathrm{T}}$ frozen, aims to adapt CLIP into the task-specific domain by using a few learnable prompts.  These extra prompts could be either concatenated to the visual~\cite{vpt}, or the textual encoder side~\cite{coop} to learn the contextual pattern tailored towards each downstream task. Specifically, $m$ visual prompts $\vp_\mathrm{V} = \{\vp_\mathrm{V}^{1},...,\vp_\mathrm{V}^{m} \} \in \R^{m \times c_{\mathrm{I}}}$, and textual prompts $\vp_\mathrm{T} = \{\vp_\mathrm{T}^{1},...,\vp_\mathrm{T}^{m} \} \in \R^{m \times c_{\mathrm{T}}}$ are concatenated to the image and textual embedding, respectively. In this way, the input image and text embedding are reformulated as $\widetilde{\tI} = \{\texttt{CLS}, \vp_\mathrm{V}, \ve_1,..., \ve_n\} \in \R^{(n+1+m) \times c_\mathrm{I}}$, and $\widetilde{\tT} = \{\vp_{\mathrm{T}}, \vt_1,..., \vt_l\} \in \R^{(l+m) \times c_{\mathrm{T}}}$. After processing these prompted image and text embeddings by $\mathcal{F}_{\mathrm{I}}$ and $\mathcal{F}_{\mathrm{T}}$,  the latent visual features $\widetilde{\mZ}_{\mathrm{I}}$ and textual features $\widetilde{\mZ}_{\mathrm{T}}$ could be obtained for further tuning CLIP as follows:
\begin{equation}\label{eq_cls}
 \mathcal{L}_{\mathrm{cls}} = - \sum\nolimits_{i=1}^{b} \hat{p}(y^{i}) \log p(y^{i}| \widetilde{\mZ}_{\mathrm{I}}^{i}, \widetilde{\mZ}_{\mathrm{T}}),
\end{equation}
where $y^{i}$ denotes the ground-truth label of sample $i$, and $\hat{p}(y^{i}) \in \{0,1\}$ is the one-hot label variable. Since PT regulates a fixed-space image ($k$-) classification problem, Eq.~(\ref{eq_cls}) only maintains the image-to-text component from the original CLIP loss. Based on this intuitive loss $\mathcal{L}_{\mathrm{cls}}$, both $\vp_{\mathrm{V}}$ and $\vp_{\mathrm{T}}$ could be optimized to improve the adaptation of CLIP for the downstream domain.

\section{Method}

\begin{figure*}[t]
\centering
\includegraphics[width=1.0\textwidth]{ 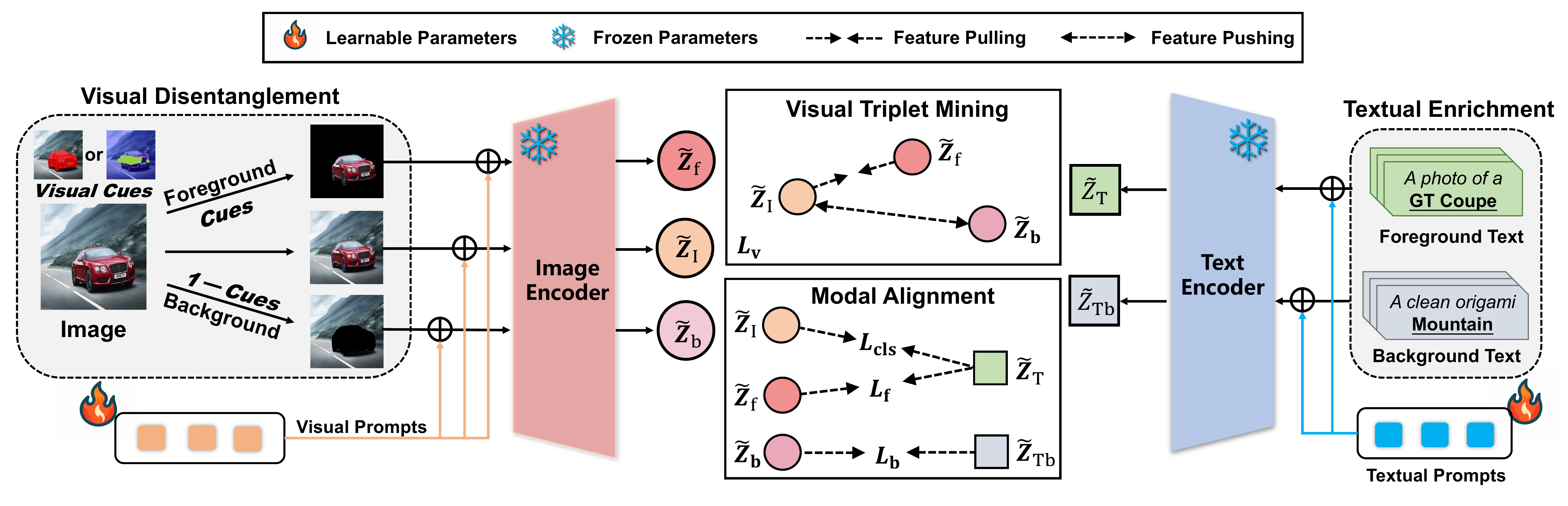}
    \vspace{-5mm}
    \caption{
The overall framework of DAPT. It mainly contains an image encoder and a text encoder. During the training, the image encoder maps the disentangled visual triplets to the feature space, which are then explicitly aligned with enriched textual features from both the foreground and background describing texts. The visual triplets are generated via either a coarse or fine-grained mask. Only the original image and foreground texts are used as input during inference. }
    \label{fig_framework}
\end{figure*}

\subsection{Information Asymmetry in Modal Alignment} \label{sec_Information Asymmetry in Modal Alignment}

As described in Eq.~(\ref{eq_cls}), PT optimizes CLIP by maximizing the similarity between the original visual and textual description. However, there exists an inherent challenge: an image, which often contains task-unrelated objects, tends to possess a stronger semantic scale compared to the single-object-oriented text description. This asymmetry in the image-text alignment, as presented in Eq. (\ref{eq_cls}), can result in what we refer to as \emph{biased attention}, where the model tends to overemphasize the background in misclassified samples. Figure~\ref{fig_illustration} shows the visualized attention of some predicted samples from two representative prompting models, i.e., CoOp~\cite{coop} and MaPLe~\cite{maple}. It is observed that compared with the rightly-predicted samples, the model's attention on the misclassified samples merely focus on the non-ROI region. In other words, these models learned from asymmetrical modal information tends to merely prioritize the context area while neglecting the oriented foreground object (additional examples can be found in Section~\ref{sec_exp}). This observation motivates us to explore a symmetrical alignment between these two modalities, which has the potential to guide the model's attention towards the ROI for those wrongly-classified samples and enhance fine-grained recognition.

\subsection{\emph{Decouple} Visual Pattern for Unbiased Recognition} \label{sec_Visual Decoupling for Context Optimization}

\noindent\textbf{Visual Disentanglement.} To effectively direct the focus of CLIP, we leverage the intriguing concept of \emph{visual cues}, which encompasses visual indicators such as points and circles that emphasize the ROI in an image. ~\citep{shtedritski2023does,yang2023set} have demonstrated that these explicit patterns can significantly enhance the fine-grained recognition capabilities of VLMs by shifting the model's attention. Motivated by this, we propose to harness this intuitive process to segregate visual information into distinct segments, paving the way for a balanced image-text alignment. Specifically, we propose segmenting the image into foreground and background components. This segmentation necessitates a semantic mask with binary values in the set $\{0,1\}$, where $1$ indicates pixels belonging to the foreground. We propose two distinct methods to generate these masks, which are:

\blackcircle{1} The first method employs a self-generated approach, deriving the mask from the visual attention map extracted from $\mathcal{F}_{\mathrm{I}}$. This process involves \emph{class activation mapping} (CAM)~\citep{cam}. CAM essentially uses a weighted combination of feature maps to effectively highlight the discriminative regions that a classifier uses to identify a specific class. This method is often used to create a coarse mask in segmentation tasks that lack pixel-level supervision~\citep{wang2018weakly,ru2022learning}. In our case, we utilize the Grad-CAM~\citep{gradcam}, a versatile CAM-based technique compatible with various network architectures, to generate semantic masks in $\mathcal{F}_{\mathrm{I}}$ by gradient information. Specifically, the Grad-CAM of an image $\tI$ \emph{w.r.t.} class $y$ is represented as $\tG_{\mathrm{I}}^{y} \in \R^{1 \times n}$, which can be computed based on
\begin{equation}\label{eq_gradcam}
   \tG_{\mathrm{I}}^{y} = \texttt{ReLU} ( \frac {1} {c_{\mathrm{I}}}
    \sum\nolimits_{i=1}^{c_{\mathrm{I}}} \frac {\partial{\texttt{sim}(\widetilde{\mZ}_{\mathrm{I}}, \widetilde{\mZ}_{\mathrm{T}}^{y})}}
    {\partial{\tA_{[i,:]}}} \circ \tA_{[i,:]}),
\end{equation}
where $\tA \in \R^{c_{\mathrm{I}} \times n}$ represents the output feature from the final transformer block in $\mathcal{F}_{\mathrm{I}}$  excluding the \texttt{CLS} token, and $\circ$ represents the Hadamard product. Subsequently, the patches are aggregated back to the original image dimensions by unfolding $n$ back to size ($h/p \times w/p$) and then interpolating to match the original image size ($h \times w$) (here $p$ refers to the patch size), yielding $\tG^{y}_{\mathrm{I}} = [g_{ij}^{y}] \in \R^{ h \times w}$. Following normalization to the range of $[0,1]$, a self-generated semantic mask can be obtained by thresholding $\tG^{y}_{\mathrm{I}}$ with a predefined $\beta$, i.e., $g_{ij}^{y} = 1 (0), g_{ij}^{y}  >  (\leq) \beta$. Here, $\beta$ helps retain the most distinctive regions in $\tG_{\mathrm{I}}$.


\blackcircle{2} 
The second approach involves generating semantic masks by using advanced and powerful segmentation tools~\citep{kirillov2023segment,seggpt}. These tools are capable of producing high-quality masks that offer detailed insights into the completeness of an object. Particularly, we turn to SEEM~\cite{seem}, an influential segmentation platform that facilitates object-level textual descriptions, to generate masks for each downstream dataset (for the textual templates, please refer to Appendix A). Formally, the mask generated by SEEM for an image $\tI$ is denoted as $\tS_{\mathrm{I}} \in \R^{h \times w}$. For convenience, we use $\tM \in \{\tG_{\mathrm{I}}, \tS_{\mathrm{I}}\}$ to represent the corresponding semantic mask of the image $\tI$.

\begin{figure}[t]
\centering
\includegraphics[width=1.0\linewidth]{  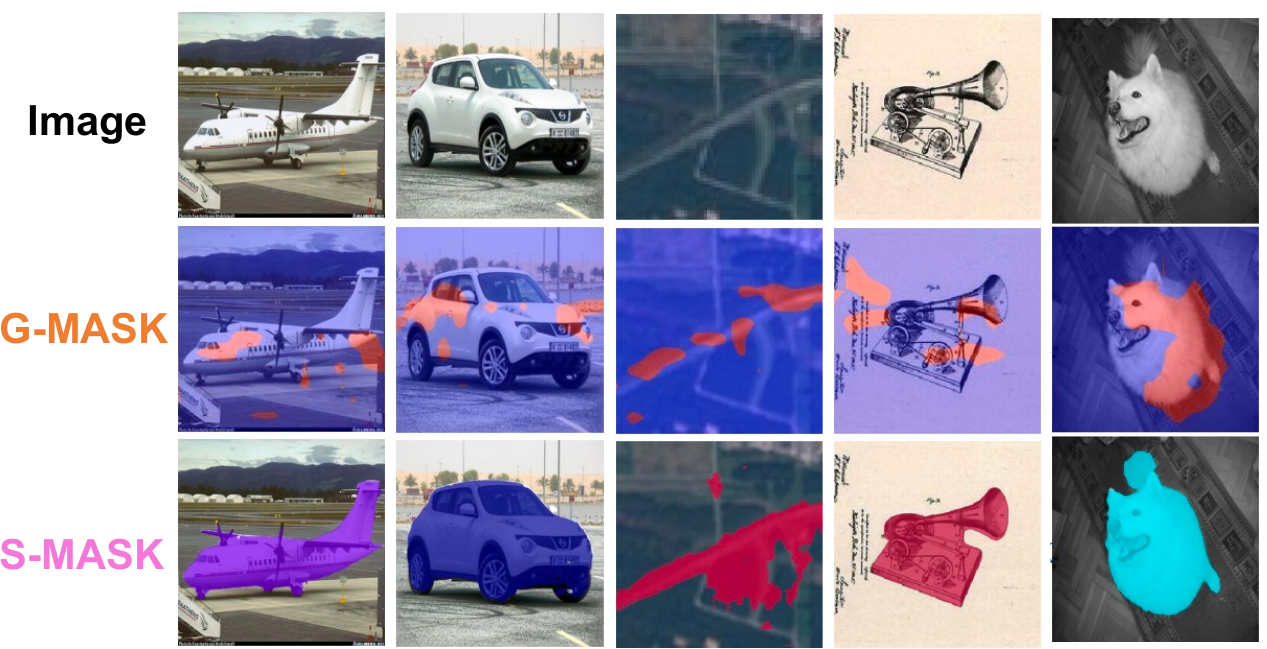}
    \vspace{-5mm}
    \caption{Illustrative samples of visual cues from DAPT-G and DAPT-S. These two methods offer a dual level of granularity in representing objectness. However, note that neither of them achieves a perfect segmentation of the query object.}
    \label{fig_visual_cues}
\end{figure}

Figure~\ref{fig_visual_cues} presents five illustrative visualized samples of these two types of semantic masks. It is evident that the Grad-CAM-based masks (G-MASK) exhibit less detailed granularity compared to the SEEM-based masks (S-MASK), which can provide nearly complete boundaries for the query object. However, it is worth noting that SEEM may not always produce a perfect 100\% accurate mask for all domains, as observed in cases like Eurosat (row 3). While S-MASK offers a better fine-grained decoupling signal, G-MASK presents a more flexible approach for PT, utilizing easily obtained and low-cost masks merely by model itself. This also highlights that our proposed method does not strictly require pixel-wise perfection in the masks for effective visual disentanglement. This observation also underscores that our proposed method does not strictly necessitate pixel-wise perfection in the masks for effective visual disentanglement.

\subsection{\emph{Align} Symmetrically for Context Optimization}\label{sec_Textual Enrichment for Context Enhancement}
To fully exploit the decoupled visual information, we propose to symmetrically align those visual patterns with the corresponding texts. Additionally, we design a pull-push term that explicitly guides the visual patterns focus more on the ROI. In this section, we shall introduce the proposed three regularizations for PT, i.e., foreground/background-text alignment, and visual triplet mining.

\noindent\textbf{Foreground/Background-Text Alignment.} Since the foreground image $\tI_{\mathrm{f}}$ shares the same semantics with the original image $\tI$ in the corresponding text, we propose to directly align the feature of $\tI_{\mathrm{f}}$ with the prompted textual feature, which could be formally expressed as
\begin{equation}\label{eq_foreground-text}
\mathcal{L}_{\mathrm{f}} = - \sum\nolimits_{i=1}^{b} \hat{p}(y^{i}) \log p(y^{i}| \widetilde{\mZ}_{\mathrm{f}}^{i}, \widetilde{\mZ}_{\mathrm{T}}).
\end{equation}

Although $\mathcal{L}_{\mathrm{f}}$ is effective in directing the focus of the model towards the query object, this may lead to the overfitting on the task-specific domain. To mitigate this, we propose to leverage the background pattern, which offers a rich context for object learning. Specifically, we propose to incorporate several textual descriptions that depict various background classes, thereby facilitating explicit alignment with $\tI_\mathrm{b}$. Suppose we introduce $k_{\mathrm{b}}$ background classes, such as \emph{ground} and \emph{land}, creating a background-based label space $\sY_{\mathrm{b}} = \{1,...,k_{\mathrm{b}}\}$. This setup generates $k_{\mathrm{b}}$ textual inputs using a predefined prompt template from~\cite{lin2023clip} (same as the zero-shot inference described in Section~\ref{sec_preliminiary}). Initially, each background image $\tI_{\mathrm{b}}^{i}$ is assigned a pseudo label, $y_{\mathrm{b}}^{i}$, determined by $y_{\mathrm{b}}^{i} = \arg\max_{j} \texttt{sim} ({\mZ}_{\mathrm{b}},{\mZ}_{\mathrm{U}}^{j}), j \in \sY_{\mathrm{b}}$. Here ${\mZ}_{\mathrm{U}}$, similar to ${\mZ}_{\mathrm{T}}$, denotes all the newly introduced background class-wise textual features. Through similar prompting operation, we shall obtain the prompted background textual features $\widetilde{\mZ}_{\mathrm{U}}$, and  set the stage for executing background-text alignment as follows:
\begin{equation}\label{eq_background-text}
 \mathcal{L}_{\mathrm{b}} = - \sum\nolimits_{i=1}^{b} \hat{p}(y_{\mathrm{b}}^{i}) \log p(y_{\mathrm{b}}^{i}| \widetilde{\mZ}_{\mathrm{b}}^{i}, \widetilde{\mZ}_{\mathrm{U}}).
\end{equation}

\noindent\textbf{Visual Triplet Mining.} 
Through $\tM$, we can generate the visual triplets for image $\tI$, namely $(\tI,\tI_{\mathrm{f}},\tI_{\mathrm{b}})$. Here, $\tI_{\mathrm{f}} = \tM \odot \tI \in \R^{3 \times h \times w}$, where $\odot$ represents the Hadamard Product, is identified as the foreground image. Conversely, $\tI_{\mathrm{b}} = (1-\tM) \odot \tI \in \R^{3 \times h \times w}$ is the background image. Our objective is to enhance the focus on the ROI by accentuating the visual pattern of $\tI_{\mathrm{f}}$. To achieve this, we propose visual triplets mining, which is implemented by using a pull-push triplet loss function~\citep{tripletloss}:
\begin{equation}\label{eq_tripletloss}
    \mathcal{L}_{\mathrm{v}} = \sum\nolimits_{i=1}^{b} \max(||\widetilde{\mZ}_{\mathrm{I}}^{i} - \widetilde{\mZ}_{\mathrm{f}}^{i}||_{1} - ||\widetilde{\mZ}_{\mathrm{I}}^{i} - \widetilde{\mZ}_{\mathrm{b}}^{i}||_{1} + \alpha, 0),
\end{equation}

where $(\widetilde{\mZ}_{\mathrm{I}}, \widetilde{\mZ}_{\mathrm{f}},\widetilde{\mZ}_{\mathrm{b}})$ represent the visual features corresponding to $(\tI,\tI_{\mathrm{f}},\tI_{\mathrm{b}})$, and $\alpha$ is a hyper-parameter that defines the minimum desired distance between $(\tI_{\mathrm{f}},\tI_{\mathrm{b}})$. Intuitively, this regularization aims to pull $\tI$ (the anchor point) closer to $\tI_{\mathrm{f}}$ (the positive point) while pushing it further from $\tI_{\mathrm{b}}$ (the negative point), thus enhancing object-level patterns in the visual representation. 

\noindent\textbf{Mask Quality.} The pull-push term $\mathcal{L}_{\mathrm{v}}$ could also relax the mask quality. Consider an extreme case where $\tM$ captures only the regions with few activated pixels. Thus, the sparsely populated $\tI_{\mathrm{f}}$, predominantly consisting of zeros, can be viewed as a masked $\tI$ with a high erasing proportion, while the background $\tI_{\mathrm{b}}$ is almost identical to $\tI$. Correspondingly, Eq.~(\ref{eq_tripletloss}) essentially becomes the maximum alignment between the original visual representation and its huge perturbed counterpart $\epsilon$, i.e., $\max(||\widetilde{\mZ}_{\mathrm{I}}^{i} - \epsilon||)$, which could serve as a form of anti-disturbance regularization for $\mathcal{L}_{\mathrm{cls}}$, implicitly aligning text with a highly-masked image. Consequently, this reveals that such regularization may not require a flawlessly accurate semantic mask, sufficing the robustness of our method. This claim will be validated in Section~\ref{sec_Effectiveness of Decoupled Pattern Learning}.

\subsection{\emph{Decouple-before-Align} Prompting Framework}\label{sec_Decouple-before-Align Prompting Framework}

Figure~\ref{fig_framework} illustrates the comprehensive architecture of our proposed DAPT. The cumulative training loss for DAPT, denoted as $\mathcal{L}_{\mathrm{all}}$, is calculated as
\begin{equation}\label{eq_final_loss}
    \mathcal{L}_{\mathrm{all}} = \gamma_\mathrm{cls}\mathcal{L}_{\mathrm{cls}} +  \gamma_\mathrm{v}\mathcal{L}_{\mathrm{v}} + 
 \gamma_\mathrm{f}\mathcal{L}_{\mathrm{f}} + 
 \gamma_\mathrm{b}\mathcal{L}_{\mathrm{b}},
\end{equation}
where $\gamma_\mathrm{cls}$, $\gamma_\mathrm{v}$, $\gamma_\mathrm{f}$, and $\gamma_\mathrm{b}$ are the hyper-parameters to balance the overall loss. Depending on the generation type of $\tM$, we designate the model as \textbf{DAPT-G} when employing Grad-CAM, and as \textbf{DAPT-S} otherwise. Since Grad-CAM could be progressively updated during the training phase, we adopt on-the-fly Grad-CAM in each epoch. During the inference stage, we only input the original testing image and the corresponding foreground-class texts for evaluation. With such an easy-to-implement loss design, our DAPT is architecture-free and can be seamlessly integrated into existing PT frameworks, which will be validated in Section~\ref{sec_Generalization_from_BtN}.

\section{Experiments}\label{sec_exp}
\subsection{Benchmark Settings}

Following~\cite{coop,maple}, we evaluate DAPT mainly based on the following settings: \textbf{I}. Few-shot classification; \textbf{II}. Data-efficient learning; \textbf{III}. Generalization from Base-to-Novel Classes.

\noindent\textbf{Datasets and Evaluation Metrics.} For all settings, we strictly follow~\cite{coop,cocoop,prograd,yao2023visual,maple} for a fair comparison by conducting the experiments on 11 datasets, i.e., ImageNet~\citep{imagenets}, Caltech101~\citep{caltech101}, OxfordPets~\citep{oxfordpets}, StanfordCars~\citep{stanford_cars}, Flowers102~\citep{flowers}, Food101~\citep{food101}, FGVCAircraft (Aircraft)~\citep{fgvc}, SUN39~\citep{sun397}, UCF101~\citep{ucf101}, DescribableTextures (DTD)~\citep{dtd}, and EuroSAT~\citep{eurosat}. For setting \textbf{III}, four ImageNet-variant datasets are additionally evaluated, which contain ImageNetV2~\citep{recht2019imagenetv2}, ImageNet-Sketch~\citep{imagnetsketch}, ImageNet-A~\citep{imagenet-a} and ImageNet-R~\citep{imagenet-r}. Unless specifically indicated, we use the prediction accuracy (\%) as the evaluation metric.

\noindent\textbf{Implementation Details.}  We, except in settings \textbf{I} and \textbf{III}, primarily use a few-shot training approach by conducting experiments with 16 randomly sampled shots per class. Adhering closely to~\cite{maple}, we apply PT to a pre-trained ViT-B/16 CLIP model. For the DAPT training, we employ a batch size of 4 and a learning rate of 0.0035, utilizing the SGD optimizer on a single NVIDIA 3090 GPU equipped with 24 GB of memory. Our experimental results are derived from the average of three trial runs. We set $m=2$ in our experiments, where $m$ refers to the number of visual prompts $\vp_\mathrm{V}$ or textual prompts ($\vp_\mathrm{T}$) that are concatenated in the image or text embedding, respectively. The language prompts for foreground and background classes are initialized using the templates “\emph{a photo of} [\texttt{FOREGROUND NAME}]” and “\emph{a clean origami} [\texttt{BACKGROUND NAME}]”, respectively. Inspired from~\cite{lin2023clip}, we employ 25 predefined background classes to constitute $\sY_{\mathrm{b}}$. The coefficients in setting \textbf{I} and \textbf{II} for $\mathcal{L}_{\mathrm{all}}$ are set as follows: $\gamma_\mathrm{cls} = 1$, $\gamma_\mathrm{v} = 0.6$, $\gamma_\mathrm{f} = 0.4$, $\gamma_\mathrm{b} = 0.1$, and $\alpha = 5.0$. For setting \textbf{III},  coefficients are adjusted to $\gamma_\mathrm{v} = 0.4$, and $\gamma_\mathrm{b} = 0.5$ (where the analysis could refer to Sec~\ref{sec_Effectiveness of Decoupled Pattern Learning}). For DAPT-G, we set $\beta = 0.5$ and incorporate the on-the-fly Grad-CAM masks in each training epoch. For the implementation of our DAPT-G and DAPT-S, we turn to the multi-modal architecture design of~\cite{maple} as the baseline. In particular, we adopted a joint prompting approach where visual prompts $\vp_\mathrm{V}$ are conditionally mapped from textual prompts $\vp_\mathrm{T}$ using a coupling (linear) function, denoted as $\vp_\mathrm{V} = \phi(\vp_\mathrm{T})$. This bridging of modalities enhances mutual synergy between visual and textual information. Additionally, the visual and textual prompts are hierarchically concatenated at various stages of transformer layers, leading to fast convergence. As illustrated in Section~\ref{sec_Decouple-before-Align Prompting Framework}, our method is not restricted to the architecture. Therefore, we also implement our concept of DAPT on current single/multi-modal-based state-of-the-art PT frameworks, e.g., CoOp~\cite{coop} and PromptKD~\cite{promptkd}, and BLIP~\cite{li2023blip}. Regarding more implementation details, we ask the readers refer to Appendix A.1.

\noindent{\textbf{Background Class.}} Following~\cite{lin2023clip}, we use 25 background classes to form the background class space, which are \emph{\{ground, land, grass, tree, building, wall, sky, lake, water, river, sea, railway, railroad, keyboard, helmet, cloud, house, mountain, ocean, road, rock, street, valley, bridge, sign.\}}

\begin{figure*}[t]
\centering
\includegraphics[width=1.0\textwidth]{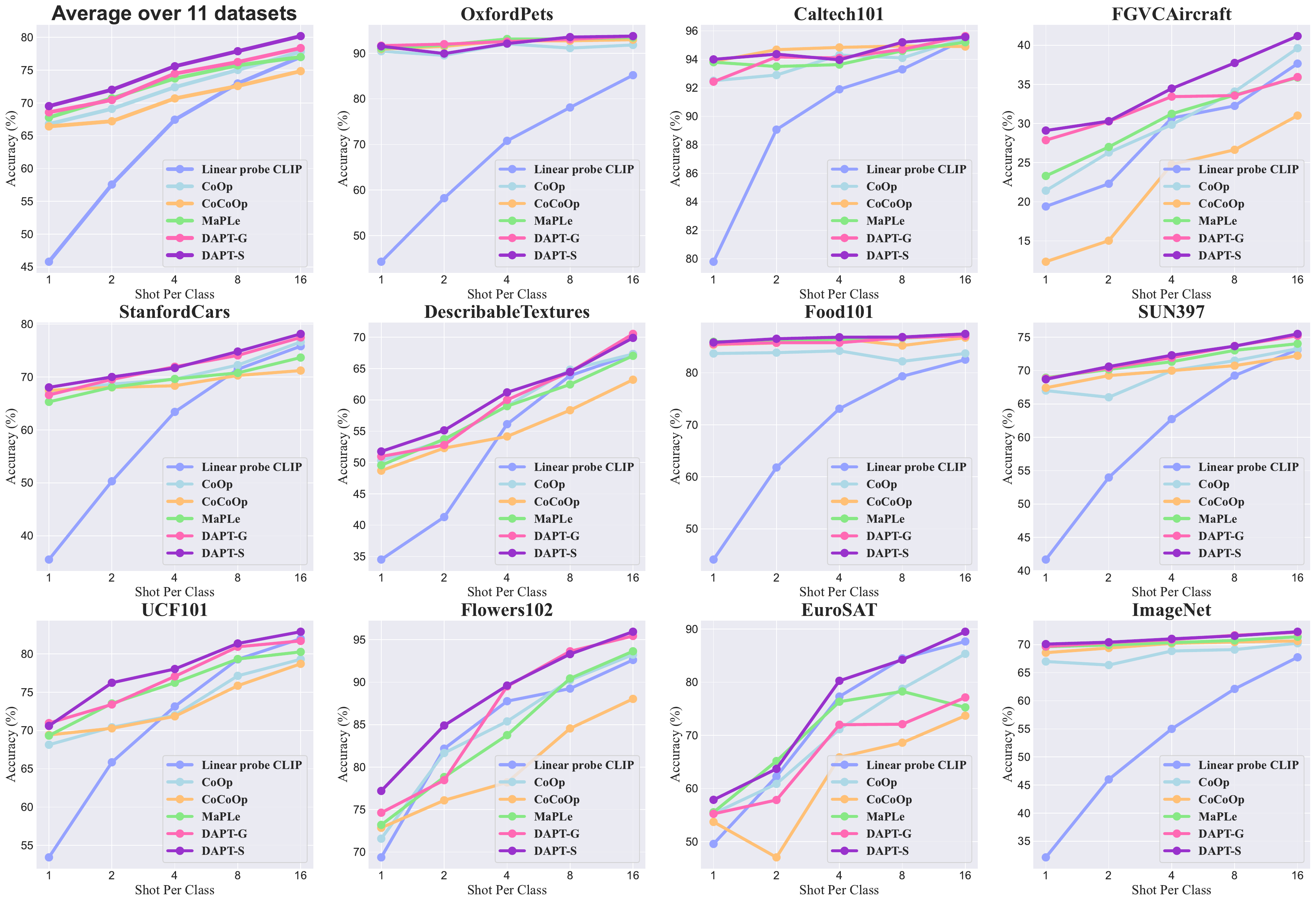}
    \vspace{-6mm}
    \caption{Performance comparison in few-shot recognition. DAPT shows remarkable performance over all existing methods, demonstrating its exceptional efficacy in domain-specific learning.}
    \label{fig_fewshot}

\end{figure*}

\subsection{Few-shot Classification (Setting I)}
\noindent\textbf{Setup.} This setting 
evaluates the effectiveness of PT under an extremely limited number of training samples. For each dataset, we follow the evaluation protocol in~\cite{clip}, where all models are trained with \{1, 2, 4, 16\} shots respectively, and then evaluated on the full test dataset. We compare DAPT against 4 methods: 1) Linear probe of CLIP, 2) CoOp~\citep{coop}, 3) CoCoOp~\citep{cocoop}, and 4) MaPLe~\citep{maple}.

\noindent\textbf{Experimental Analysis.} 
Figure~\ref{fig_fewshot} showcases a comprehensive comparison of these five methods. Impressively, DAPT outperforms all its competitors. Notably, DAPT-S consistently delivers a significant performance boost (an average of \textbf{+1.94\%}) in accuracy across varying shot scenarios. DAPT-G, while slightly less effective than DAPT-S due to its mask granularity, still stands out among the methods, underscoring the potent impact of local \emph{visual disentanglement} in enhancing PT. Notably, DAPT-G occasionally performs slightly better than DAPT-S, which typically could be attributed to the \emph{training randomness or instability} in few-shot settings~\cite{coop,maple}. These results robustly affirm the effectiveness and dominance of DAPT in mastering task-specific patterns. We also observe a consistent competitive performance gap between DAPT-S and other methods on several non-natural benchmarks, such as EuroSAT. We attribute this marginal improvement of our method to the absence of task priors for mask decoupling, with further discussion available in Appendix D.

\begin{figure}[htbp]
\centering
    \includegraphics[width=0.90\linewidth]{ 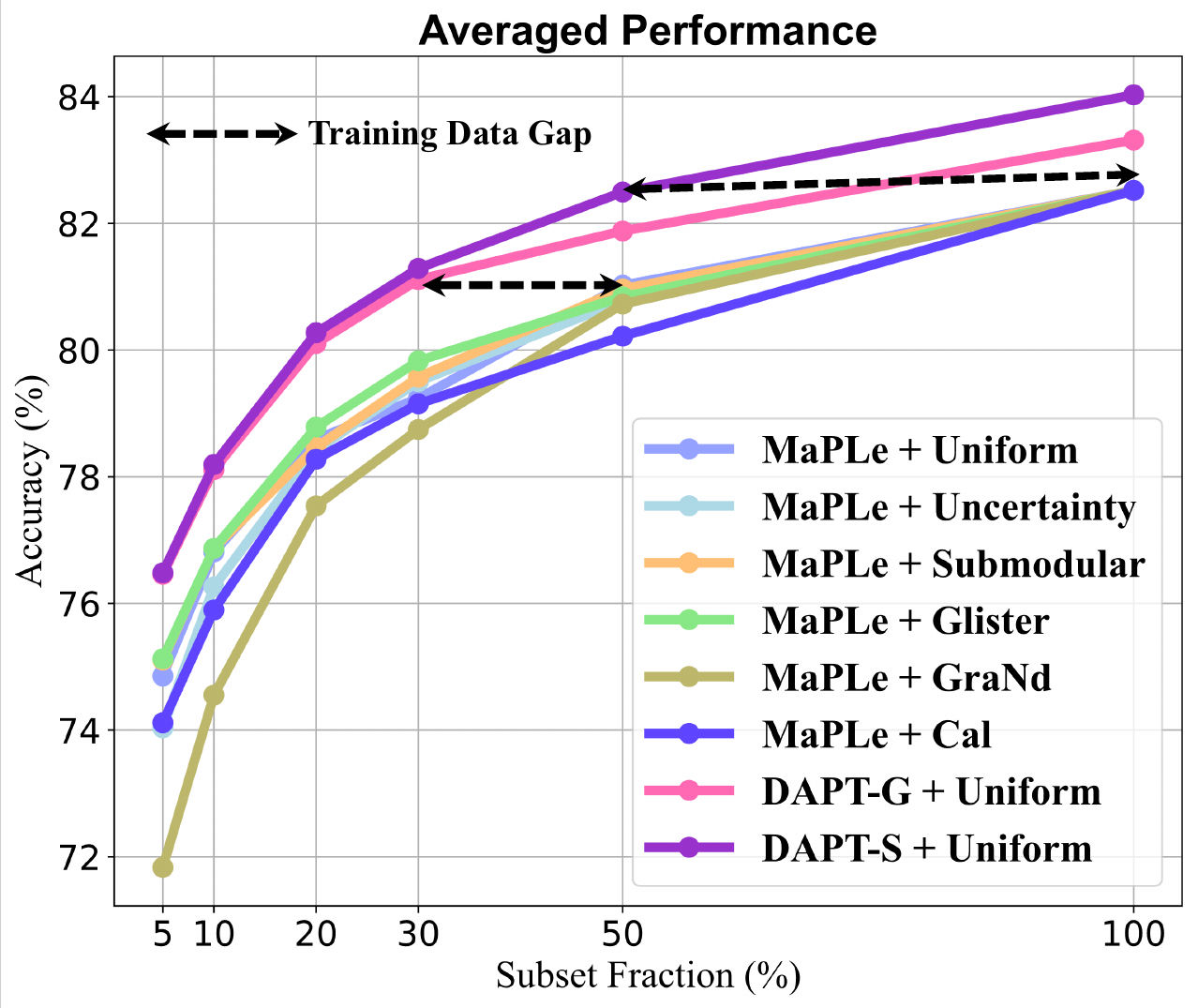}
    \vspace{-3mm}
    \caption{Performance comparison in Data-efficient setting. As marked by the black arrow, DAPT exhibits strong data-efficient performance which could even save 50\% training data.}
    \label{fig_dataefficient}

\end{figure}

\subsection{Data-efficient Learning (Setting II)}

\noindent\textbf{Setup.} As reported in~\cite{coop, maple}, it is evident that model performance is strongly influenced by the number of training shots (as also demonstrated in Figure~\ref{fig_fewshot}), underscoring the significance of training data volume. This has piqued our interest in exploring the upper-bound performance of DAPT and its potential for achieving data-efficient learning. We aim to ascertain whether DAPT can maintain a promising performance level when trained on a smaller subset of the entire training dataset. Different from few-shot learning, our focus is on optimizing performance across the entirety of the training data, rather than solely relying on randomly selected few-shot samples. To this end, we follow~\cite{guo2022deepcore}, and implement 6 data selection methods on MaPLe~\cite{maple} as the baselines for comparison. These methods are designed to select the samples beneficial to the model most. These methods are 1) Random Selection (the same as the default setting), 2) Submodular~\cite{submodular}, 3) Entropy Uncertainty~\cite{coleman2019selection}, 4) Glister~\citep{killamsetty2021glister}, 5) GraNd~\citep{grand}, and 6) Cal~\citep{cal}. We merely use the Random Selection strategy for DAPT-S/G. We set the training data subsets with fractions of \{5\%, 10\%, 20\%, 30\%, 50\%, 100\%\}. For the evaluation data, to facilitate comparisons on a unified scale, we report the averaged results across 10 of the fine-grained datasets, excluding ImageNet due to its overwhelming scale. Regarding more implementation of these methods, we ask readers to refer to Appendix A.2.

\noindent\textbf{Experimental Analysis.} As shown in Figure~\ref{fig_dataefficient} (additional details are in Appendix A.2), DAPT-S/G exhibits remarkable data-efficient capabilities, surpassing other selective-based methods by achieving comparable performance to MaPLe by merely using \textbf{50\%} of the training data. This efficiency is indicated by the black arrow,
highlighting how DAPT-S/G reduces the required data volume across different subset fractions. Specifically, DAPT-G and DAPT-S models achieve \textbf{81.63\%} and \textbf{82.51\%} in performance, respectively, yielding an average accuracy improvement of \textbf{0.92}\% and \textbf{1.77}\% compared to other methods. These results underscore the resource-efficient nature of DAPT. Interestingly, we observe that the performance gap between DAPT-S and DAPT-G widens as the volume of training data increases, emphasizing the positive impact of fine-grained masks.

\begin{table}[h]
\centering
\caption{Performance comparison in Domain-specific Base-to-Novel. Here the harmonic mean (\%) is also reported.}
\label{tab_base_to_novel}
\resizebox{0.95\linewidth}{!}{
\begin{tabular}{l|c|c|c|c}
\toprule[1.0pt]
Method  & Year & Base & Novel   &  HM   \\ 

\midrule
CLIP~\citep{clip}  & ICLM21 & 69.34 & 74.22 & 71.70 \\
CoOp~\citep{coop}  & IJCV22 & 82.69 & 63.22 & 71.66 \\
CoCoOp~\cite{cocoop} & CVPR22 & 80.47 & 71.69 & 75.83 \\

MaPLe~\cite{maple} & CVPR23 & 82.28 & 75.14 & 78.55 \\ 
ProGrad~\cite{prograd} & ICCV23 & 82.48 & 70.75 & 76.16 \\ 
KgCoOp~\cite{yao2023visual} & CVPR23 & 80.73 & 73.60 & 77.00 \\ 
\midrule
DAPT-G (ours) & - & 83.13 & 74.14 & 78.38 \\ 
DAPT-S (ours) & - & \textbf{83.95} & \textbf{75.23} & \textbf{79.35} \\ 
\bottomrule[1pt]
\toprule[1.0pt]
PromptSRC~\cite{selfregulating} & ICCV23 & 84.26 & 76.10 & 79.97 \\ 
PromptKD~\cite{promptkd} & CVPR24 & 86.96 & 80.73 & 83.73 \\
\midrule
DAPT\textsubscript{+PromptSRC} & - & 86.11 & 77.05 & 81.32 \\ 
DAPT\text{\scriptsize+PromptKD}  & - & \textbf{88.05} & \textbf{81.14} & \textbf{84.45}
 \\ 
\bottomrule[1pt]
\end{tabular}
}
\end{table}

\subsection{Generalization from Base-to-Novel (Setting III)}\label{sec_Generalization_from_BtN}
\subsubsection{Domain-specific Base-to-Novel }


\begin{table*}[t]
\centering
\caption{Cross-Data Evaluation on 10 fine-grained classification datasets. DAPT also reaches comparable performance in both target and imagenet-based source domains, indicating a promising out-of-domain generalization capability.}
\resizebox{1.0\textwidth}{!}{
\begin{tabular}{cc cc c cccccccccc}
\toprule[1pt]
\multicolumn{2}{c}{\multirow{2}{*}{Method}} & \multicolumn{2}{c}{\textbf{Source}} & & \multicolumn{10}{c}{\textbf{Target}} \\ 
\cline{3-4} \cline{6-15} 
\vspace{-3mm}\\
& & \multicolumn{2}{c}{ImageNet} & &Flowers102 & Food101 & AirCraft  & EuroSAT & OxfordPets & StanfordCars  & SUN397 & DTD & UCF101 & Caltech101    \\ 
\midrule
\multicolumn{2}{c}{CoOp} & \multicolumn{2}{c}{71.51}  &  & 68.71 & 85.30 & 18.47 & 46.39 & 89.14 &  64.51  &  64.15 & 41.92 & 66.55 & 93.70 \\ 
\multicolumn{2}{c}{CoCoOp} & \multicolumn{2}{c}{71.02}   &  & 71.88 & 86.06 & 22.94 & 45.37 & 90.14 & 65.32 & 67.36 & 45.73 & 68.21 & \textbf{94.43}\\ 
\multicolumn{2}{c}{MaPLe} & \multicolumn{2}{c}{70.72}   &  & 72.23 & \textbf{86.20} & 24.74 & \textbf{48.06} & 90.49 & \textbf{65.57} & 67.01 & 46.49 & \textbf{68.69} & 93.53\\  

\midrule
\multicolumn{2}{c}{DAPT-G} & \multicolumn{2}{c}{71.63}   &  & 71.78 & 85.21 & 23.01   & 46.16 & \textbf{90.92} & 65.44 & 66.39 & 45.21 & 68.62 & 93.41 \\ 
\multicolumn{2}{c}{DAPT-S} & \multicolumn{2}{c}{\textbf{71.71}}   &  & \textbf{72.52} & 85.96  & \textbf{24.92} & 46.12 & 90.24 & 65.52 & \textbf{67.53} & \textbf{47.31} & 68.36 & 93.81\\ 

\bottomrule[1pt]
\toprule[1pt]
\multicolumn{2}{c}{PromptSRC} & \multicolumn{2}{c}{71.27}   &  & 70.25 & 86.15 & 23.90 & 45.50 & 89.14 & 64.51 & 67.36 & 41.92 & 66.55 & 93.70\\ 
\multicolumn{2}{c}{PromptKD} & \multicolumn{2}{c}{78.12}   &  & 75.33 & 88.84 & 26.24 & 63.74 & 90.14 & 65.32 & 67.10 & 45.73 & \textbf{68.21} & 93.61\\ 
\midrule
\multicolumn{2}{c}{DAPT\textsubscript{+PromptSRC}} & \multicolumn{2}{c}{73.45}   &  & 71.39 & 87.96  & 25.11 & 47.06 & 89.91 & 65.31 & \textbf{67.98}  & 42.35 &67.43 & \textbf{93.89}\\
\multicolumn{2}{c}{DAPT\textsubscript{+PromptKD}} & \multicolumn{2}{c}{\textbf{78.91}}   &  & \textbf{75.87} & \textbf{89.12}  & \textbf{27.14} & \textbf{64.36} & \textbf{90.87} & \textbf{65.83} & 67.66 & \textbf{46.35} & 67.92 & 93.72\\
\bottomrule[1pt]
\end{tabular}
}
\label{tab_crossdata}
\end{table*}
\begin{table}[h]
\centering
\caption{Cross-Data Evaluation on 4 ImageNet-based datasets. Notably, DAPT also reaches superior performance in both target and source domains for these ImageNet-varaiant benchmarks, indicating a promising out-of-domain generalization capability.}

\resizebox{1.0\linewidth}{!}{
\begin{tabular}{cc cc c cccc}
\toprule[1pt]
\multicolumn{2}{c}{\multirow{2}{*}{Method}} & \multicolumn{2}{c}{\textbf{Source}} & & \multicolumn{4}{c}{\textbf{Target}} \\ 
\cline{3-4} \cline{6-9} 
\vspace{-3mm}\\
& & \multicolumn{2}{c}{ImageNet} & &-V2& -S & -A & -R    \\ 
\midrule
\multicolumn{2}{c}{CoOp} & \multicolumn{2}{c}{71.51}  &  &  64.20 & 47.99 & 49.71 & 75.21  \\ 
\multicolumn{2}{c}{CoCoOp} & \multicolumn{2}{c}{71.51}  &  &  64.07 & 48.75 & 50.63 &76.18  \\ 
\multicolumn{2}{c}{MaPLe} & \multicolumn{2}{c}{71.51}  &  &64.07 &49.15& \textbf{50.90}& 76.98  \\ 
\midrule
\multicolumn{2}{c}{DAPT-G} & \multicolumn{2}{c}{71.63}  & & 64.51 &48.82& 47.97 &76.82  \\ 
\multicolumn{2}{c}{DAPT-S} & \multicolumn{2}{c}{\textbf{71.71}}  &&  \textbf{64.43} & \textbf{49.43}&49.41 &\textbf{77.22}  \\ 
\bottomrule[1pt]
\toprule[1pt]
\multicolumn{2}{c}{PromptSRC} & \multicolumn{2}{c}{71.27}  &&   64.35 &49.55 &50.90 &77.80  \\ 
\multicolumn{2}{c}{PromptKD} & \multicolumn{2}{c}{78.12}  && 69.77 &58.72 &\textbf{70.36} &87.01  \\ 
\midrule
\multicolumn{2}{c}{DAPT\textsubscript{+PromptSRC}} & \multicolumn{2}{c}{73.45} &&  64.81 & 49.98 &51.32 &77.14  \\ 
\multicolumn{2}{c}{DAPT\textsubscript{+PromptKD}} & \multicolumn{2}{c}{\textbf{78.91}}  &&  \textbf{70.11} & \textbf{59.07} &69.73 &\textbf{87.52} \\ 
\bottomrule[1pt]
\end{tabular}
}
\label{tab_crossdata_imgNet}
\end{table}

\noindent\textbf{Setup.} This setting aims to assess the model's ability to generalize from seen classes to unseen classes in task-specific domains, which is also the main target for most PT frameworks. Following~\citep{prograd,maple}, we partition the dataset into equal subsets containing seen and unseen classes. Subsequently, we train the models using the seen classes and conduct evaluations on both the seen and unseen class subsets. Additionally, we report the \emph{harmonic mean} (HM) for each dataset. Here we compare our DAPT with 8 PT prevailing frameworks. For the implementation of our DAPT-G and DAPT-S, we turn to the multi-modal architecture design of~\cite{maple} as the baseline. As illustrated in Section~\ref{sec_Decouple-before-Align Prompting Framework}, our method is not restricted to the architecture. Therefore, we here implement our concept of DAPT on two PT frameworks, i.e., PromptSRC~\cite{selfregulating} and PropmptKD~\cite{promptkd}, both of which achieve the state-of-the-art performance. Note that PromptKD is a two-stage teacher-student framework that distilled from first-stage-pre-trained PromptSRC. DAPT (-S)+PromptSRC is implemented through adding the designed regularization, i.e., $\mathcal{L}_{\mathrm{v}} + \mathcal{L}_{\mathrm{f}} +\mathcal{L}_{\mathrm{b}}$
, to the optimization of PromptRC. DAPT(-S) + PromptKD is implemented through using DAPT+PromptSRC as the teacher, thereby guiding PromptKD to a better student. This conclusion aligns with the findings in~\cite{selfregulating}.

\noindent\textbf{Experimental Analysis.} Table~\ref{tab_base_to_novel} presents a comparative analysis of the accuracy and HM for eight distinct methods across 11 datasets. DAPT consistently outperforms others in recognizing both base and novel class images, yielding an average accuracy improvement of $\textbf{+1.67\%}$ and an increase of $\textbf{+0.80\%}$ in HM. Besides, despite DAPT-G's commendable performance in base class recognition, it exhibits less proficiency in novel class identification, which can be attributed to its coarse disentanglement. In contrast, DAPT-S not only captures the fundamental representation more effectively but also demonstrates equal or superior capability in learning novel representations, showcasing its potent generalization. When incorporating DAPT-S as a plug-and-play module, it is observed an consistent base-and-novel improvement on both PromptSRC and PromptKD, with an average increase of \textbf{+1.03\%} across this task. In this way, our DAPT with PromptKD achieves the SOTA performance with a leading \textbf{84.45\%} HM. Overall, DAPT achieves a win-win situation between base and novel class recognition.

\subsubsection{Cross-Data Base-to-Novel }

\noindent\textbf{Setup.} This configuration assesses the out-of-domain generalization capabilities of models pre-trained on ImageNet, which are then evaluated on various downstream datasets in a zero-shot manner. Following the paradigm in \cite{cocoop,prograd,maple}, we train DAPT-S by using 16-shot examples from each of the 1000 classes in ImageNet, and then evaluate the model performance on other prevailing benchmarks.

\noindent\textbf{Experimental Analysis.} Table~\ref{tab_crossdata_imgNet} \& \ref{tab_crossdata} delineate the cross-dataset generalizability of several methods across 14 distinct datasets, including 10 fine-grained classification and 4 ImageNet-based recognition benchmarks. As shown in Table~\ref{tab_crossdata}, notably, DAPT-S/G outshines its counterparts in terms of domain transfer capabilities, where DAPT-S has achieving favorable recognition in source domain (\textbf{71.71\%} accuracy) while strengthening powerful target-domain recognition. Besides, as a plug-and-play module, our DAPT has achieved the best recognition performance in both source and target domain, accompanied by a comprehensive improvement against all downstream benchmarks with an average of \textbf{+1.18\%} (\textbf{+0.69\%}) performance elation on PromptSRC (PromptKD). Addtionally, as shown in Table~\ref{tab_crossdata_imgNet}, our DAPT is also effective on ImageNet-variant benchmarks, and both the baseline DAPT and DAPT+PromptSRC has reached SOTA performance across 3 of 4 ImageNet-based datasets.  The outcomes of these experiments provide further corroboration that DAPT has successfully attained a harmonious balance between source and target domain performances, thereby showcasing its robustness in mitigating domain shifts.

\subsection{Multi-object Recognition}
\noindent{\textbf{Setup.}} Previous efforts~\cite{coop,cocoop,maple,promptkd} have merely been evaluated among those fine-grained single-object classification problems. Here we, from a more practical perspective, explore the potential application of these methods in multi-object scenarios with 20 classes in total.  Since VOC12 is a 20-class classification dataset, we replace cross-entropy loss  $\mathcal{L}_{\mathrm{cls}}$ as \emph{multi-label soft-margin loss}. For the generated masks in DAPT-G and DAPT-S, we combine all the segmented foreground objects into one co-foreground as $\tI_{\mathrm{f}}$, thereby forming the same visual triplets for performing visual disentanglement. The other training parameters are aligned with the original setting. The evaluation metric is \emph{mean Average Precision} (mAP) (\%).

\noindent\textbf{Multi-object Recognition.} As shown in Table~\ref{tab_voc12}, firstly, it could be seen an overall high recognition ability delivered by these PT frameworks, all of which achieve about 90\% mAP with given 16-shot samples. Secondly, it is observed that DAPT also shows superior performance in both few-shot and base-to-novel cases on VOC12, achieving a leading role with an average of $\textbf{+3.13}$ elation compared to other methods. Particularly, it is observed that DAPT shall bring higher performance improvement than those fine-grained datasets, which could attribute to more common natural objects in VOC12. Overall, the results above validate the effectiveness of our proposed DAPT in addressing multi-object scenarios.

\begin{table}[t]
\caption{Performance on VOC12. The metric is mAP (\%)}
\centering
\resizebox{1.0\linewidth}{!}{
\setlength{\tabcolsep}{1mm}{
\begin{tabular}{ c  cc c cc}
\toprule[1pt]
 \multirow{2}{*}{Method}  & \multicolumn{2}{c}{Few-Shot} & & \multicolumn{2}{c}{Base-to-Novel} \\
 
 \cline{2-3} \cline{5-6}  
 \vspace{-3mm}\\
   & \multicolumn{2}{c}{1 / 4 / 16-shot Accuracy}  & &\multicolumn{2}{c}{Base / Novel / HM} \\
\midrule
CoOp  & \multicolumn{2}{c}{66.52 / 78.34 / 88.12} & &\multicolumn{2}{c}{93.34 / 71.42 / 80.92} \\
CoCoOp  & \multicolumn{2}{c}{70.65 / 82.15 / 90.44} & &\multicolumn{2}{c}{92.73 / 78.68 / 85.13} \\
MaPLe  & \multicolumn{2}{c}{74.56 / 89.42 / 93.41} & &\multicolumn{2}{c}{95.01 / 83.40 / 88.83} \\
\midrule
DAPT-G  & \multicolumn{2}{c}{\textbf{75.77 / 91.86 / 95.22}} & &\multicolumn{2}{c}{\textbf{96.02 / 84.12 / 89.73}} \\
DAPT-S  & \multicolumn{2}{c}{\textbf{77.21 / 92.71 / 96.88}} & &\multicolumn{2}{c}{\textbf{96.88 / 85.32 / 90.73}} \\
\bottomrule[1pt]
\end{tabular}
}
}
\label{tab_voc12}
\end{table}

\subsection{Ablation Studies}\label{sec_Effectiveness of Decoupled Pattern Learning}
In this section, we discuss the effectiveness of the designed modules in DAPT via a broad range of in-depth experiments, including the effectiveness of the loss modules and the corresponding regularized weights, the mask quality, the masking strategy, and the computational efficiency. To further verify the superiority of our DAPT, we also evaluate the performance of our DAPT on multi-object real-world classification problem.  We, unless specifically indicated, adopt the averaged results of DAPT-S with 16-shot StanfordCars and ImageNet for all ablation studies. (\textcolor{orange}{\ding{72}} More implemented experimental results, including \emph{complex textual case}, and \emph{advanced trial on BLIP} can be found in Appendix.)


\begin{table}[htbp]
\caption{Effectiveness of loss items on DAPT.}
\centering
\resizebox{1.0\linewidth}{!}{
\setlength{\tabcolsep}{1mm}{
\begin{tabular}{ c c c c  cc c cc}
\toprule[1pt]
 \multirow{2}{*}{Baseline}  & \multirow{2}{*}{$\mathcal{L}_{\mathrm{v}}$}  & \multirow{2}{*}{$\mathcal{L}_{\mathrm{f}}$}  & \multirow{2}{*}{$\mathcal{L}_{\mathrm{b}}$}  & \multicolumn{2}{c}{Few-Shot} & & \multicolumn{2}{c}{Base-to-Novel} \\
 \cline{5-6} \cline{8-9}  
 \vspace{-3mm}\\
 &  &  &  & \multicolumn{2}{c}{1 / 4 / 16-shot Accuracy}  & &\multicolumn{2}{c}{Base / Novel / HM} \\
\midrule
\ding{52}  &  &  &  & \multicolumn{2}{c}{67.51 / 70.25 / 72.71} & &\multicolumn{2}{c}{74.81 / 72.27 / 73.47} \\



\ding{52}   &  \ding{52}  & &   & \multicolumn{2}{c}{\textcolor{tgreen}{+0.94} / \textcolor{tgreen}{+1.50} / \textcolor{tgreen}{+1.44}} & & \multicolumn{2}{c}{\textcolor{tgreen}{+1.81} / \textcolor{tred}{-1.21} / \textcolor{tgreen}{+0.43}}   \\
\ding{52}   &  &  \ding{52}  &  & \multicolumn{2}{c}{\textcolor{tgreen}{+0.69} / \textcolor{tgreen}{+1.33} / \textcolor{tgreen}{+1.23}}  & &
\multicolumn{2}{c}{\textcolor{tgreen}{+1.46} / \textcolor{tred}{-1.28} / \textcolor{tgreen}{+0.05}}
\\
\ding{52}   &   &  &  \ding{52}  & \multicolumn{2}{c}{\textcolor{tgreen}{+0.39} / \textcolor{tgreen}{+0.96} / \textcolor{tgreen}{+1.05}} & &  \multicolumn{2}{c}{\textcolor{tgreen}{+0.23} / \textcolor{tgreen}{+1.60} / \textcolor{tgreen}{+0.93}}\\

\ding{52}   &  \ding{52}  & \ding{52}  &  & \multicolumn{2}{c}{\textcolor{tgreen}{+1.48} / \textcolor{tgreen}{+2.12} / \textcolor{tgreen}{+2.42}} & &  \multicolumn{2}{c}{\textcolor{tgreen}{+3.31} / \textcolor{tred}{-2.24} / \textcolor{tgreen}{+0.48}}\\

\ding{52}   &  \ding{52}  &   & \ding{52} & \multicolumn{2}{c}{\textcolor{tgreen}{+1.17} / \textcolor{tgreen}{+1.61} / \textcolor{tgreen}{+1.87}} & &  \multicolumn{2}{c}{\textcolor{tgreen}{+2.12} / \textcolor{tred}{-0.04} / \textcolor{tgreen}{+1.04}}\\

\ding{52}   &    & \ding{52} & \ding{52}  &  \multicolumn{2}{c}{\textcolor{tgreen}{+1.02} / \textcolor{tgreen}{+1.76} / \textcolor{tgreen}{+1.50}} &&\multicolumn{2}{c}{\textcolor{tgreen}{+1.38} / \textcolor{tgreen}{+0.34} / \textcolor{tgreen}{+0.87}}  \\

\ding{52}   &  \ding{52}   &  \ding{52}   &  \ding{52}  &  
\multicolumn{2}{c}{\textbf{69.07} / \textbf{72.37} / \textbf{75.22}} & &\multicolumn{2}{c}{\textbf{77.18} / \textbf{72.33} / \textbf{74.67}}

\\
\bottomrule[1pt]
\end{tabular}
}
}
\label{tab_effectiveloss}
\end{table}

\noindent\textbf{Individual Loss Regularization.} Our first investigation centers on the impact of $\mathcal{L}_{\mathrm{v}}$, $\mathcal{L}_{\mathrm{f}}$ and $\mathcal{L}_{\mathrm{b}}$. As shown in Table~\ref{tab_effectiveloss}, we observe that both $\mathcal{L}_{\mathrm{v}}$ and $\mathcal{L}_{\mathrm{f}}$ significantly bolster performance in few-shot scenarios. Particularly, a simple addition with $\mathcal{L}_{\mathrm{v}}$ boosts the baseline model with an average of $\textbf{+1.29}\%$ elation on few-shot learning. Compared to $\mathcal{L}_{\mathrm{v}}$, the foreground-text alignment also shows a comparable performance improvements, with achieving an average of $\textbf{+1.08}\%$ accuracy. Both $\mathcal{L}_{\mathrm{v}}$ and $\mathcal{L}_{\mathrm{f}}$ tend to guide the model to focus more on learning foreground objects, thereby intensifying the model's emphasis on in-domain object recognition alongside $\mathcal{L}_{\mathrm{cls}}$ during the training. While this results in significant improvements in in-domain performance, it also contributes to increased overfitting  (as been noted in~\cite{coop,maple}), which reasonably yields degraded novel-class recognition performance (-1.21\% / -1.28\%). Based on this, the combination of $\mathcal{L}_{\mathrm{v}}+\mathcal{L}_{\mathrm{f}}$ could significantly enhance the model's in-domain recognition ability, while causing a huge degraded out-of-domain performance as well ($-2.24\%$). On the contrary, $\mathcal{L}_\mathrm{b}$ targets on learning the newly-introduced background patterns. During training, $\mathcal{L}_\mathrm{b}$ helps reduce the emphasis on in-domain objects dictated by $\mathcal{L}_{\mathrm{cls}}$ by providing generalized contextual knowledge that extends beyond foreground patterns. This approach alleviates base-class overfitting and improves the model's generalization performance, enhancing out-of-domain foreground recognition ($\textbf{+1.60\%}$) through a better understanding of the background context. As a result, the integration of these regularizations fosters a balanced advancement between base-novel generalization, sufficiently validating their effectiveness.

\begin{figure}[t]
\centering
\begin{minipage}{1.0\linewidth}
    \includegraphics[width=1.0\linewidth]{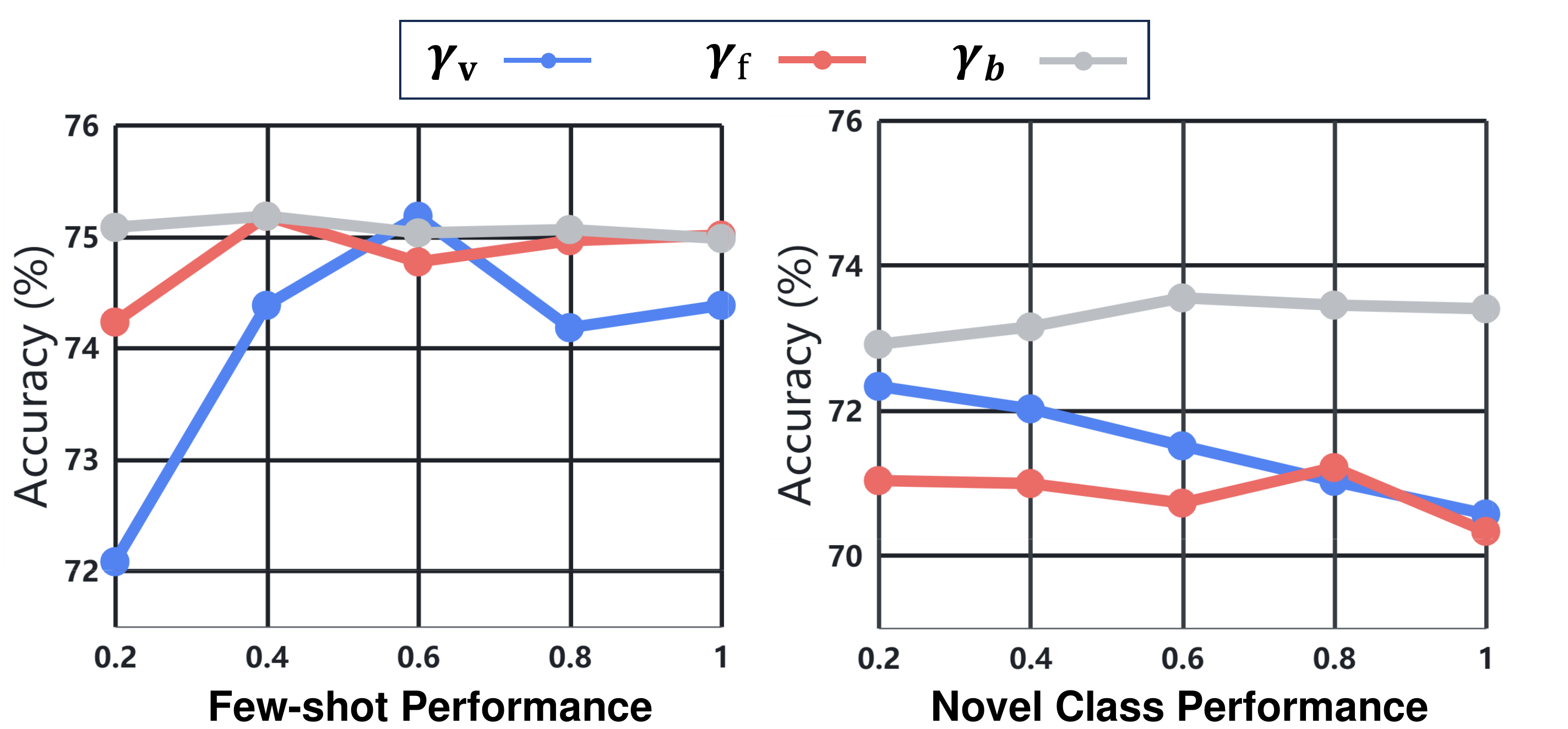}
    \vspace{-5mm}
    \caption{The effectiveness of the weight on each loss item.}
    \label{fig_para}
\end{minipage}
\end{figure}

\noindent\textbf{Regularized Weights.} To investigate the robustness of the designed loss items, Figure~\ref{fig_para} presents the impact of varying the loss weights, $\gamma_\mathrm{v}, \gamma_\mathrm{f}$, and $\gamma_\mathrm{b}$, on the model performance. We alter one weight of each loss at a time, holding the others constant, to isolate the effects of each regularization term. As demonstrated in this Figure, despite of varying combinations, our DAPT could holistically surpass the baselines with only a minor fluctuation of \textbf{0.54\%} averaged on different hyper-parameters. Specifically, in the case of few-shot learning, $\mathcal{L}_{\mathrm{v}}$ exhibits greater sensitivity to its corresponding weight compared to $\mathcal{L}_{\mathrm{f}}$, whereas $\mathcal{L}_{\mathrm{b}}$ shows a remarkable insensitivity to parameter changes, contributing to a consistently stable performance. For novel class recognition, an increase in $\gamma_\mathrm{b}$, regardless of insignificant performance fluctuation, enhances the model's generalization capabilities (demonstrating the robustness of DAPT to the weight.). In contrast,  $\mathcal{L}_{\mathrm{v}}$ and $\mathcal{L}_{\mathrm{f}}$ tend to overfit the task-specific domain.  This also explains that \emph{we adjust the corresponding weights in setting \textbf{III}}, leading to an optimal numerical performance between base and novel class recognition. According to the above analysis, we could conclude that all of these hyper-parameters of DAPT could be easily optimized to achieve superior improvements without costly trials, thereby verifying the robustness of DAPT. 

\begin{table}[t]
\centering
\caption{Ablations on the separate term in $\mathcal{L}_{\mathrm{v}}$. }
\label{tab_loss_v}
\resizebox{1.0\linewidth}{!}{
\setlength{\tabcolsep}{1mm}{
\begin{tabular}{ c  cc c cc}
\toprule[1pt]
 \multirow{2}{*}{Loss items}  & \multicolumn{2}{c}{Few-Shot} & & \multicolumn{2}{c}{Base-to-Novel} \\
 
 \cline{2-3} \cline{5-6}  
 \vspace{-3mm}\\
   & \multicolumn{2}{c}{1 / 4 / 16-shot Accuracy}  & &\multicolumn{2}{c}{Base / Novel / HM} \\
\midrule
Baseline   & \multicolumn{2}{c}{67.51 / 70.25 / 72.71} & &\multicolumn{2}{c}{74.81 / 72.27 / 73.47} \\
\midrule
+Foreground-Positive  & \multicolumn{2}{c}{\textcolor{tgreen}{+0.67} / \textcolor{tgreen}{+1.18} / \textcolor{tgreen}{+1.03}} & &\multicolumn{2}{c}{\textcolor{tgreen}{+1.52} / \textcolor{tred}{-0.79} / \textcolor{tgreen}{+0.11}} \\
+Background-Negative & \multicolumn{2}{c}{\textcolor{tgreen}{+0.44} / \textcolor{tgreen}{+0.62} / \textcolor{tgreen}{+0.77}} & &\multicolumn{2}{c}{\textcolor{tgreen}{+0.25} / \textcolor{tred}{-0.19} / \textcolor{tgreen}{+0.13}} \ \\
+$\mathcal{L}_{\mathrm{v}}$  & \multicolumn{2}{c}{\textbf{\textcolor{tgreen}{+0.94} / \textcolor{tgreen}{+1,0} / \textcolor{tgreen}{+1.44}}} & &\multicolumn{2}{c}{\textbf{\textcolor{tgreen}{+1.81} / \textcolor{tred}{-1.21} / \textcolor{tgreen}{+0.43}}}
 \\
\bottomrule[1pt]
\end{tabular}
}
}
\end{table}

\noindent\textbf{Visual Triplet Components.} Recall that the introduced visual triplet mining, i.e., $\mathcal{L}_{\mathrm{v}}$, is essentially comprised of two parts, i) aligning the original visual pattern close to the foreground; ii) pushing the former one away from the background. Therefore, it is crucial to display how the separation of foreground and background as positive and negative samples improves over the original triplet loss. To this end, we conduct this ablations through adding $||\widetilde{\mZ}_{\mathrm{I}}^{i} - \widetilde{\mZ}_{\mathrm{f}}^{i}||_{1}$
 (Foreground-Positive), and $-||\widetilde{\mZ}_{\mathrm{I}}^{i} - \widetilde{\mZ}_{\mathrm{b}}^{i}||_{1}$
 (Background-Negative), respectively. As shown in Table~\ref{tab_loss_v} it is seen that both the foreground-background visual components, though slightly inferior to 
, contribute to a certain level of improvement. Specifically, they tend to enhance in-domain knowledge acquisition but adversely affect novel recognition, with the background cases showing a less severe diminish.

\noindent\textbf{Mask Quality.} 
In Section~\ref{sec_Visual Decoupling for Context Optimization}, we highlighted that our method \emph{does not necessitate an excessively precise mask}, as evidenced by the performance comparison between DAPT-G and DAPT-S. To provide a more tangible verification, we first conduct an illustrative experiment where we randomly erased foreground regions of the masks in DAPT-S and DAPT-G at varying ratios. It is emphasized that the self-generated Grad-CAM, accompanied by weak-label-triggered noise, could somewhat mimic real-world masks~\cite{cam,shen2023survey}. As shown in Table~\ref{tab_erase}, the 4 / 16-shot recognition performance, while experiencing a significant drop beyond an erasing rate of 0.7, remains at a reasonable level within the erasing rate range of 0.1 to 0.5 for both DAPT-G and DAPT-S, further validating the mask-free robustness of DAPT.
\begin{table}[t]
\centering
\caption{The influence of mask quality under erasing strategy on 4/16 Few-Shot task. Here the greater erasing rates represents more grid-like destroyed regions to the foreground mask. }
\resizebox{1.0\linewidth}{!}{
\begin{tabular}{ c  |c|c|c|c}
\toprule[1pt]
  Erasing Rate & 0.1 & 0.3 & 0.5  & 0.7  \\
  \hline
    DAPT-G  & 70.96 / 73.22 & 69.86 / 72.41 & 69.11 / 71.97 & 68.98 / 71.35 \\
  \hline
  DAPT-S  & 72.11 / 75.22& 72.03 / 75.05 & 71.94 / 74.87 & 71.34 / 73.97 \\

\bottomrule[1pt]
\end{tabular}
}
\label{tab_erase}
\end{table}

\begin{table}[t]
\centering
\caption{Performance of DAPT with mask from different segmentation generation methods. }
\label{tab_ap_different_mask_method}
\resizebox{1.0\linewidth}{!}{
\setlength{\tabcolsep}{1mm}{
\begin{tabular}{ c  cc c cc}
\toprule[1pt]
 \multirow{2}{*}{Method}  & \multicolumn{2}{c}{Few-Shot} & & \multicolumn{2}{c}{Base-to-Novel} \\
 
 \cline{2-3} \cline{5-6}  
 \vspace{-3mm}\\
   & \multicolumn{2}{c}{1 / 4 / 16-shot Accuracy}  & &\multicolumn{2}{c}{Base / Novel / HM} \\
\midrule
DAPT-G   & \multicolumn{2}{c}{68.12 / 71.14 / 74.17} & &\multicolumn{2}{c}{76.15 / 71.04 / 73.51} \\
DAPT\textsubscript{+CLIP-ES}  & \multicolumn{2}{c}{68.48 / 71.31 / 74.36} & &\multicolumn{2}{c}{76.54 / 71.32 / 73.84} \\
DAPT\textsubscript{+FreeSeg}  & \multicolumn{2}{c}{68.77 / 71.72 / 74.90} & &\multicolumn{2}{c}{76.86 / 71.94 / 74.32} \\
DAPT-S   & \multicolumn{2}{c}{\textbf{69.07 / 72.37 / 75.22}} & &\multicolumn{2}{c}{\textbf{77.18 / 72.33 / 74.67}} \\
\bottomrule[1pt]
\end{tabular}
}
}
\end{table}

To further support the above argument, we further adopt two segmentation frameworks, i.e., CLIP-ES~\cite{lin2023clip} and FreeSeg ~\cite{qin2023freeseg}. The former one, like Grad-CAM, generates the semantic masks by merely using the image-text from CLIP, while the latter one is a universal segmentation framework trained based on large-scale segmentation benchmarks. According to their reported performance in VOC12, we shall obtain that the performance ranking (mIoU) of the segmenting ability of these models is Grad-CAM  < CLIP-ES  < FreeSeg < SEEM. As shown in Table~\ref{tab_ap_different_mask_method}, our method with these four different segmenting models could achieve an overall promising performance elation,  which further validates the robustness of our method towards the mask quality. We will add this analysis. Reasonably, the performance of DAPT increases with better mask quality, but such a modest growth reveals the mask-tolerance of our method.  

\begin{table}[t]
\caption{The influence of mask generated by Gaussian Blurring (GB). Following~\cite{yang2023fine}, here GB is implemented with $(5,9)$ kernel size and $(0.1,1.0)$ sigma.}
\centering
\resizebox{1.0\linewidth}{!}{
\setlength{\tabcolsep}{1mm}{
\begin{tabular}{ c  cc c cc}
\toprule[1pt]
 \multirow{2}{*}{Method}  & \multicolumn{2}{c}{Few-Shot} & & \multicolumn{2}{c}{Base-to-Novel} \\
 
 \cline{2-3} \cline{5-6}  
 \vspace{-3mm}\\
   & \multicolumn{2}{c}{1 / 4 / 16-shot Accuracy}  & &\multicolumn{2}{c}{Base / Novel / HM} \\
\midrule
Baseline   & \multicolumn{2}{c}{67.51 / 70.25 / 72.71} & &\multicolumn{2}{c}{	74.81 / 72.27 / 73.47} \\
DAPT+GB  & \multicolumn{2}{c}{68.35 / 71.77 / 74.09} & &\multicolumn{2}{c}{76.11 / 72.31 / 74.16} \\
DAPT+HF   & \multicolumn{2}{c}{\textbf{69.07 / 72.37 / 75.22}} & &\multicolumn{2}{c}{\textbf{77.18 / 72.33 / 74.67}} \\
\bottomrule[1pt]
\end{tabular}
}
}
\label{tab_erase_gb}
\end{table}

\begin{table}[t]
\centering
\caption{Effectiveness of background classes. Here Only $\mathcal{L}_\mathrm{b}$ is adopted within our DAPT.}
\resizebox{1.0\linewidth}{!}{
\setlength{\tabcolsep}{1mm}{
\begin{tabular}{ c  cc c cc}
\toprule[1pt]
 \multirow{2}{*}{\makecell{Background \\ Numbers}}  & \multicolumn{2}{c}{Few-Shot} & & \multicolumn{2}{c}{Base-to-Novel} \\
 
 \cline{2-3} \cline{5-6}  
 \vspace{-3mm}\\
   & \multicolumn{2}{c}{1 / 4 / 16-shot Accuracy}  & &\multicolumn{2}{c}{Base / Novel / HM} \\
\midrule
5 & \multicolumn{2}{c}{ 66.01 / 70.87 / 72.96} & &\multicolumn{2}{c}{74.89 / 72.17 / 73.47 } \\
10 & \multicolumn{2}{c}{ 66.42 / 70.98 / 73.16} & &\multicolumn{2}{c}{74.94 / 72.67 / 73.76 } \\
15 & \multicolumn{2}{c}{ 67.63 / 71.07 / 73.45} & &\multicolumn{2}{c}{75.01 / 73.11 / 74.07 } \\
25 & \multicolumn{2}{c}{ \textbf{67.90 / 71.21 / 73.76}} & &\multicolumn{2}{c}{\textbf{75.04 / 73.87 / 74.40}} \\
\bottomrule[1pt]
\end{tabular}
}
}

\label{tab_backgroundclass}
\end{table}

\noindent\textbf{Masking Strategy.} DAPT adopts an intuitive 0-1 mask for disentangling the visual patterns. Except for such a Hard filling (HF) manner for masking, we here, inspired from~\cite{yang2023fine}, conduct another masking strategy on \emph{visual disentanglement} by turning to Gaussian Blurring, which shall better preserve the overall visual relationship between foreground and background in images. In this way, the generated blurring mask could be treated as a soft label of 0-1 mask. Compared to HF, GB is supposed to work in a broader way since it considers the dark object/scene cases. However, it brings less visual difference for the disentangled triplets. As shown in Table~\ref{tab_erase_gb}, despite the observed baseline-level improvement, GB generally shows slightly inferior performance compared to HF. In comparison to GB, HF more effectively leverages our visual triplet mechanism by creating a more significant foreground-background differentiation, thereby enhancing the evaluated fine-grained pattern recognition. Based on this, we shall conclude that while GB may function in a more versatile manner for DAPT, it tends to show comparatively inferior performance to HF in most natural domains.

\noindent\textbf{Background Classes} To enable background-text alignment, we introduce 25 background classes to create a background space. We conducted an investigation into the effectiveness of varying the number of background classes in our proposed method. Table~\ref{tab_backgroundclass} presents the few-shot and base-to-Novel performance of our method under different numbers of background classes. It is evident that the few-shot performance of our model, while not outstanding, exhibits an increase as the number of background classes rises. This suggests the minimal effectiveness of the background class number for task-specific learning. However, this number significantly impacts the learning of novel classes, leading to substantial improvements as the background space expands. This is reasonable because enriching the background space helps the model align with more contextual information. These experimental results highlight the importance of background learning in enhancing out-of-domain generalizations.

\begin{table}[h]
\centering
\caption{Performance of single-modal prompting architectures. }
\label{tab_single_modal}
\resizebox{1.0\linewidth}{!}{
\setlength{\tabcolsep}{1mm}{
\begin{tabular}{ c  cc c cc}
\toprule[1pt]
 \multirow{2}{*}{Method}  & \multicolumn{2}{c}{Few-Shot} & & \multicolumn{2}{c}{Base-to-Novel} \\
 
 \cline{2-3} \cline{5-6}  
 \vspace{-3mm}\\
   & \multicolumn{2}{c}{1 / 4 / 16-shot Accuracy}  & &\multicolumn{2}{c}{Base / Novel / HM} \\
\midrule
CoOp   & \multicolumn{2}{c}{66.35 / 68.56 / 70.41} & &\multicolumn{2}{c}{76.40 / 64.14 / 69.74} \\
+DAPT  & \multicolumn{2}{c}{\textbf{\textcolor{tgreen}{\textbf{+1.12}} / \textcolor{tgreen}{\textbf{+2.36}} / \textcolor{tgreen}{\textbf{+2.85}}}} & &\multicolumn{2}{c}{\textbf{\textcolor{tgreen}{\textbf{+1.35}} / \textcolor{tgreen}{\textbf{+0.92}} / \textcolor{tgreen}{\textbf{+1.11}}} }\\
\midrule
VPT  & \multicolumn{2}{c}{	64.79 / 67.72 / 68.81} & &\multicolumn{2}{c}{73.94 / 62.77 / 67.89} \\
+DAPT  & \multicolumn{2}{c}{\textbf{\textcolor{tgreen}{+2.34} / \textcolor{tgreen}{+2.14} / \textcolor{tgreen}{+3.67}}} & &\multicolumn{2}{c}{\textbf{\textcolor{tgreen}{+2.25} / \textcolor{tgreen}{+0.23} / \textcolor{tgreen}{+1.08}
}} \\
\bottomrule[1pt]
\end{tabular}
}
}
\end{table}

\begin{table}[t]
\centering
\caption{Performance of DAPT with different ViT backbones. }
\label{tab_ap_different_backbones}
\resizebox{1.0\linewidth}{!}{
\setlength{\tabcolsep}{1mm}{
\begin{tabular}{ c  cc c cc}
\toprule[1pt]
 \multirow{2}{*}{Method}  & \multicolumn{2}{c}{Few-Shot} & & \multicolumn{2}{c}{Base-to-Novel} \\
 
 \cline{2-3} \cline{5-6}  
 \vspace{-3mm}\\
   & \multicolumn{2}{c}{1 / 4 / 16-shot Accuracy}  & &\multicolumn{2}{c}{Base / Novel / HM} \\
\midrule
DAPT-G\textsubscript{+ViT-B}   & \multicolumn{2}{c}{68.12 / 71.14 / 74.17} & &\multicolumn{2}{c}{76.15 / 71.04 / 73.51} \\
DAPT-G\textsubscript{+ViT-L}   & \multicolumn{2}{c}{\textcolor{tgreen}{+1.34} / \textcolor{tgreen}{+1.29} / \textcolor{tgreen}{+1.94}} & &\multicolumn{2}{c}{\textcolor{tgreen}{+2.14} / \textcolor{tgreen}{+0.98} / \textcolor{tgreen}{+1.51}} \\
\midrule
DAPT-S\textsubscript{+ViT-B}   & \multicolumn{2}{c}{69.07 / 72.37 / 75.22} & &\multicolumn{2}{c}{77.18 / 72.33 / 74.67} \\
DAPT-S\textsubscript{+ViT-L}   & \multicolumn{2}{c}{\textcolor{tgreen}{+1.75} / \textcolor{tgreen}{+1.49} / \textcolor{tgreen}{+1.65}} & &\multicolumn{2}{c}{\textcolor{tgreen}{+2.57} / \textcolor{tgreen}{+1.12} / \textcolor{tgreen}{+1.79}} \\
\bottomrule[1pt]
\end{tabular}
}
}
\end{table}

\begin{figure*}[t]
\centering
    \includegraphics[width=1.0\linewidth]{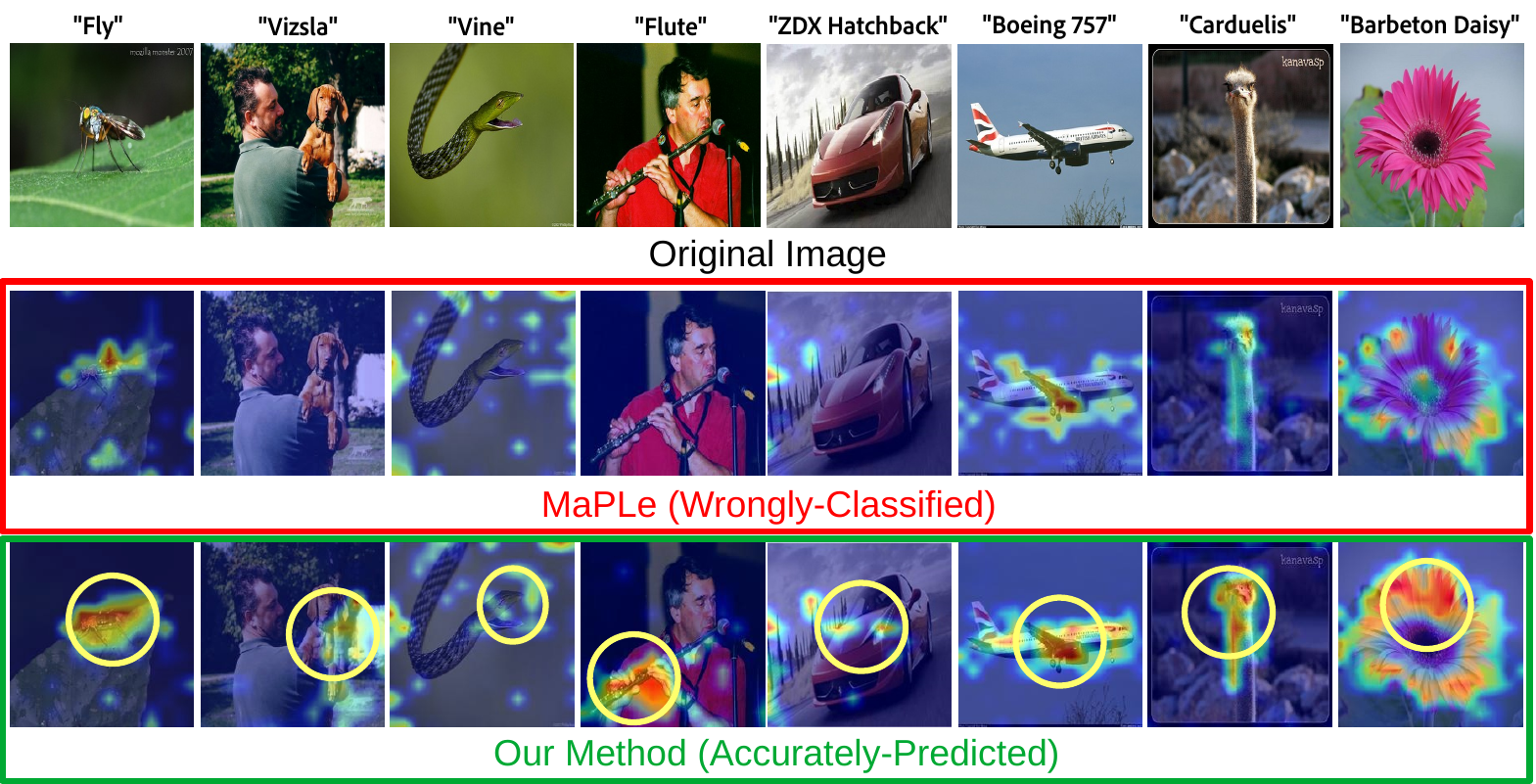}
    \vspace{-5mm}
    \caption{The visualized results between MaPLe and DAPT-S. These images are selected from ImageNet, StanfordCars, Flowers102, Aircraft, and OxfordPets. Compared to MaPLe, our method could help drastically help model focus on the query ROI part to correct the misclassified samples. The yellow circle highlights object-oriented attention.}
    \label{fig_visual}

\end{figure*}

\begin{table}[t]
\centering
\caption{Time computation (in minutes) on ImageNet. The batch size during inference is set to $128$ across all methods. *Note that for DAPT-S, the reported time is only for the $1st$ epoch since all annotation is once-for-all pre-finished by SEEM.}
\resizebox{1.0\linewidth}{!}{
\begin{tabular}{ c  cc c | c}
\toprule[1pt]
  \multirow{2}{*}{Method} & \multicolumn{2}{c}{Training Time} & & \multirow{2}{*}{\makecell{Testing Time \\ (50000 samples)}} \\
 \cline{2-3}   
     & \multicolumn{2}{c}{1 / 4 / 16-shot Accuracy}  & &  \\
\midrule
CoOp & \multicolumn{2}{c}{ 0.17 / 0.41 / 1.79 } & &{3.96} \\
MaPLe & \multicolumn{2}{c}{ 0.17 / 0.41 / 1.79 } & &{3.96} \\
\midrule
DAPT-G & \multicolumn{2}{c}{ 0.37 / 0.74 / 2.33 } & &{3.98} \\
DAPT-S* & \multicolumn{2}{c}{ 	0.33 / 0.98 / 3.61 } & &{3.98} \\
\bottomrule[1pt]
\end{tabular}
}
\label{tab_compute_time}
\end{table}

\noindent\textbf{Single-modal-prompted Adaptation.} Our baseline DAPT is built and evaluated on multi-modal prompting architectures, which shall have better performance than single-modal-based prompting methods. Due to the multi-modal nature of VLM, the co-exist image-text encoders both contribute towards efficiently aligning the VL modalities. Correspondingly, as also validated in~\cite{maple,promptkd}, optimizing the single-modal prompt shall not sufficiently model the adaptations needed for another modality. However, we would like to claim that our DAPT could be effectively employed on the image/text methods with adaptable modification. To verify this, we instantiate DAPT on two single-modal-prompted frameworks, i.e., VPT (image)~\cite{vpt} and CoOp (text)~\cite{coop}, across few-shot learning and base-to-novel tasks. For VPT, we keep the original loss with merely optimizing visual prompts. For CoOp, as lacking visual prompts update, we merely adopt the foreground/background-text ($\mathcal{L}_{\mathrm{f}} + \mathcal{L}_{\mathrm{b}}$) alignment for prompting the whole architecture. As presented in Table~\ref{tab_single_modal}, both VPT and CoOp show marked improvements with our designed modules, highlighting the consistent effectiveness of DAPT in single-modal-prompting case. 

\noindent\textbf{Model Scaling.} To evaluate the scaling ability of our method, we have conducted the experiments using DAPT with two commonly used ViT backbones: ViT-B/16 (baseline) and the more powerful ViT-L/14. As illustrated in Table~\ref{tab_ap_different_backbones}, upgrading the backbone leads to better enhancements in both few-shot and base-to-novel recognition tasks, resulting in average performance increases of $\textbf{+1.57\%}$ and $\textbf{+1.39}$ (HM), respectively. These observed improvements further validate the promising scalability of our method in terms of both in-domain and out-of-domain generalization. 

\noindent\textbf{DAPT-G \emph{vs.} DAPT-S.} Since \emph{visual disentanglement} is only applied for training, thus no disentanglement technique is required during the test. Therefore, there should be no extra inference computational costs about DAPT. Regarding the training efficiency, although decoupling visual patterns brings extra pre-processing complexity, we clarify that such a cost is reasonably acceptable against other vanilla methods. Table~\ref{tab_compute_time} shows the per-epoch-training and inference time on ImageNet. Clearly, DAPT-S merely takes an average of \textbf{0.21 seconds per 40 images} before the training (batch size as 40 for SEEM, 3.5G memory occupation). In other words, the 16-shot 1000-class experiments merely bring once-for-all preprocessing costs about \textbf{1.4 minutes} for the whole training. For DAPT-G, the on-the-fly Grad-CAM generation simply takes an additional average \textbf{5.13 seconds per epoch} during the overall training phase. Overall, the above numerical training-inference results shall demonstrate a manageable and comparable level of computational efficiency for DAPT. 

Notably, while DAPT-S generally demonstrates better performance, \textbf{DAPT-G offers a more cost-effective and versatile training manner for PT.} DAPT-G utilizes self-generated masks from the VLM itself, eliminating the need for external segmentation tools and significantly reducing the annotation costs for visual decoupling. In this way, this on-the-fly generated representation can be seamlessly integrated into PT. Clearly, with more samples involved in the training process, DAPT-G exhibits improved training efficiency compared to DAPT-S, which incurs additional pre-processing costs for mask generation. Furthermore, the investigation of DAPT-G provides a novel perspective on leveraging weakly generated signals to enhance PT by relaxing the strict pixel-level mask granularity required for effective improvement (as shown in Table 7 \& 8), thereby validating the broader applicability of our DAPT. Thus, we believe that this use of self-knowledge represents a valuable exploration for PT.

\noindent
In conclusion, we recommend DAPT-S as the top choice, when we have auxiliary good segmentation model available. However, without any external segmentation tools, DAPT-G can be mostly considered, since the on-the-fly generation of Grad-CAM can be easily acquired and integrated into PT.

\noindent\textbf{Visualized Analysis.} In Section~\ref{sec_Information Asymmetry in Modal Alignment}, we discussed the concept of \emph{information asymmetry} leading to biased attention, where the model demonstrates an inadequate focus on the query texts. The motivation behind \emph{visual disentanglement} is to direct the model's attention towards the query object. As depicted in Figure~\ref{fig_visual}, our approach drives CLIP towards the right recognition by concentrating more on the ROI. Surprisingly, this method can also globally activate or enhance the attention towards the previously overlooked foreground portion, further validating its ability to improve multi-domain recognition. However, it is also found our DAPT may still exhibit high-level attention to partial background in some cases, e.g., the vine snake, the roadside trees next to the car, and the sky behind the airplane, which is also reasonable since background serves an important context for fine-grained recognition~\cite{sagawa2019distributionally,moayeri2022comprehensive,xiao2020noise}. Such a context-preserved capability shall attribute our loss design. Instead of completely removing the background as a negative element, the valuable background-aware prior is also reflected through the original image/background-text alignment in DAPT, i.e., $\mathcal{L}_{\mathrm{b}}$. Therefore, these visualized results also valiate the preservation of valid context recognition of our method.


\section{Conclusion and Future Work}
This paper has illuminated a previously overlooked issue in PT for VLMs: The conventional asymmetrical alignment of the prompted image-text pairs can result in \emph{biased attention} from CLIP, diverging from the query ROI for the misclassified samples. To address this challenge, we investigate the use of \emph{visual cues} that explicitly decouples the image into foreground and background patterns, and then correspondingly enhance the textual representations to achieve symmetrical modal alignment, encompassing foreground-text and background-text. Through both quantitative and qualitative experiments, we have showcased the effectiveness and superiority of this straightforward \emph{decouple-before-align} concept across various in-domain and out-of-domain tasks. This adjustment directs the attention of CLIP toward object-oriented patterns in an unbiased manner. Furthermore, this work highlights that this PT mechanism for VLMs, unlike previous rigid fine-tuning approaches against global parameters, can be accomplished through a simple yet explicit visual signal. We hope this opens avenues for further exploration in PT.

Despite our method achieving state-of-the-art performance, DAPT struggles to effectively address the challenge of distinguishing between base and novel classes, particularly in non-natural benchmarks like DTD and EuroSAT, where there is a significant performance disparity. Additionally, our method is only evaluated in single/multi-object classification problems, its application may be limited to other tasks, such as VQA.  Finally, our approach, along with other pipelines, primarily focuses on the two modality-based (image-text) architectures, suggesting that a broader range of Multi-modal VLMs, such as video, should also be considered.

\ifCLASSOPTIONcompsoc
  \section*{Acknowledgments}
\else
  \section*{Acknowledgment}
\fi
This work is supported by the National Key R\&D Program of China (No. 2022ZD0160703), National Natural Science Foundation of China (No. 62306178), STCSM (No. 22DZ2229005), 111 plan (No. BP0719010), and Beijing Natural Science Foundation (L252036).

%
\ifCLASSOPTIONcaptionsoff
  \newpage
\fi



%
{
\bibliographystyle{IEEEtran}
\bibliography{example_paper}
}




%

\begin{IEEEbiography}
[{\includegraphics[width=1in,height=1.25in,clip,keepaspectratio]{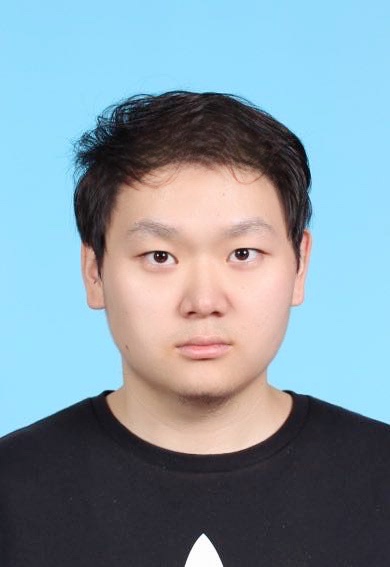}}]{Fei Zhang}
is pursuing the Ph.D. degree in Shanghai Jiao Tong University, Shanghai, China. He is also with Shanghai Innovation Institute. Prior to that, he obtained his B.S. degree in automation from Northwestern Polytechnical University, and M.S. degree in control science and engineering from Shanghai Jiao Tong University. His
research interests include visual recognition, image segmentation, multi-modal foundation model, and active learning.
\end{IEEEbiography}

\begin{IEEEbiography}
[{\includegraphics[width=1in,height=1.25in,clip,keepaspectratio]{  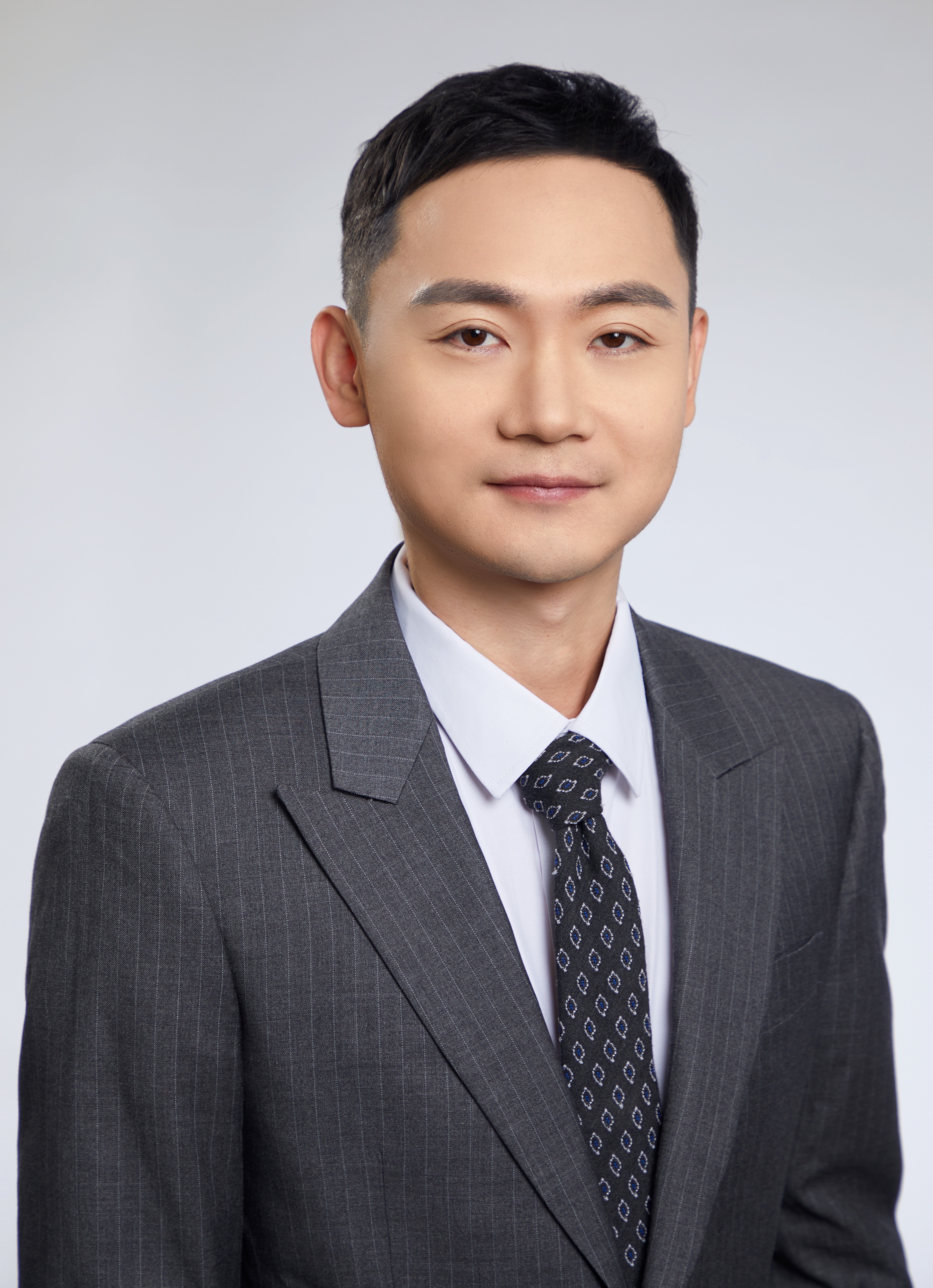}}]{Tianfei Zhou}
is currently a Professor with Department of Computer
Science, Beijing Institute of Technology, China. Prior to that, he was a
research fellow with Computer Vision Lab, ETH Zurich, Switzerland. He
obtained his Ph.D. degree from Beijing Institute of Technology in 2017.
His current research interests are mainly in the areas of computer vision,
medical image analysis and machine learning. He was the recipient of
MICCAI MEDIA Best Paper Award in 2022.
\end{IEEEbiography}


\begin{IEEEbiography}[{\includegraphics[width=1in,height=1.25in,clip,keepaspectratio]{  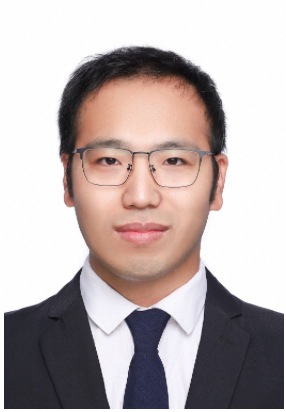}}]{Jiangchao Yao}
is an Assistant Professor of
Shanghai Jiao Tong University, Shanghai, China.
He received the B.S. degree in information engineering from South China University of Technology, Guangzhou, China, in 2013. He got a dual
Ph.D. degree under the supervision of Ya Zhang
in Shanghai Jiao Tong University and Ivor W.
Tsang in University of Technology Sydney. His
research interests include deep representation
learning and robust machine learning.
\end{IEEEbiography}

\begin{IEEEbiography}
[{\includegraphics[width=1in,height=1.25in,clip,keepaspectratio]{  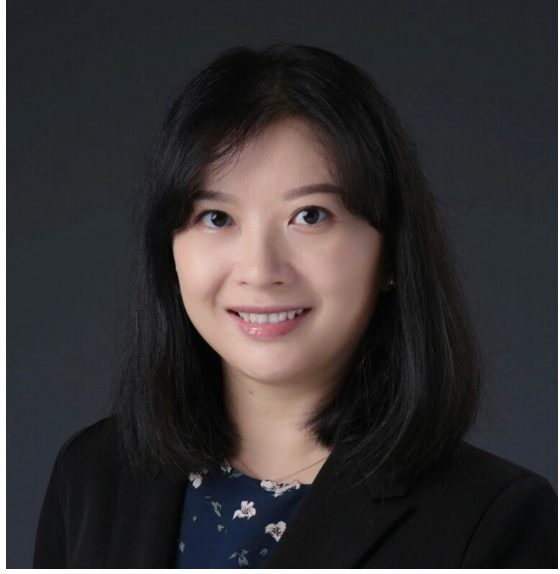}}]{Ya Zhang} 
is currently a Professor in School of Artificial Intelligence, Shanghai Jiao
Tong University. Her research interest is mainly
on machine learning and data mining with applications to multimedia information retrieval, social network analysis, and intelligent information
system. Prof. Zhang holds a PhD degree in Information Sciences and Technology from Pennsylvania State University and a Bachelor’s degree
from Tsinghua University in China. Prof. Zhang
published more than 70 refereed papers in prestigious international
conferences and journals including TPAMI, TIP, TNNLS, ICDM, CVPR,
ICCV, ECCV, and ECML. 
She is appointed as the Chief Expert for the project ’Research of
Key Technologies and Demonstration for Digital Media Self-organizing’
under the 863 program by Ministry of science and technology of China.
\end{IEEEbiography}

\begin{IEEEbiography}[{\includegraphics[width=1in,height=1.25in,clip,keepaspectratio]{  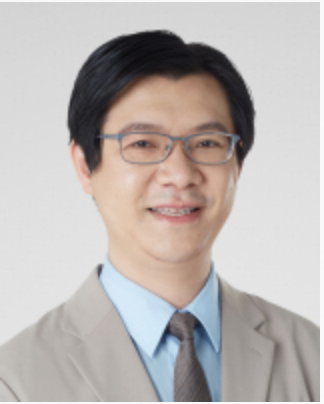}}]{Ivor W. Tsang}
is a Professor of Artificial Intelligence, at University of Technology Sydney
(UTS). He is also the Research Director of the
UTS Flagship Research Centre for Artificial Intelligence (CAI). His research focuses on transfer
learning, feature selection, big data analytics for
data with extremely high dimensions in features,
samples and labels, and their applications to
computer vision and pattern recognition. He has
more than 190 research papers published in top-tier journal and conference papers. According to
Google Scholar, he has more than 10,000 citations and his H-index is
56. In 2009, Prof. Tsang was conferred the 2008 Natural Science Award
(Class II) by Ministry of Education, China, which recognized his contributions to kernel methods. In 2013, Prof. Tsang received his prestigious
Australian Research Council Future Fellowship for his research regarding Machine Learning on Big Data. In 2019, he received the International
Consortium of Chinese Mathematicians Best Paper Award in recognition
of his work ”Towards ultrahigh dimensional feature selection for big
data”, published in Journal of Machine Learning Research. In addition,
he had received the prestigious IEEE Transactions on Neural Networks
Outstanding 2004 Paper Award in 2007, the 2014 IEEE Transactions
on Multimedia Prize Paper Award, and a number of best paper awards
and honors from reputable international conferences, including the Best
Student Paper Award at CVPR 2010.
\end{IEEEbiography}

\begin{IEEEbiography}[{\includegraphics[width=1in,height=1.25in,clip,keepaspectratio]{  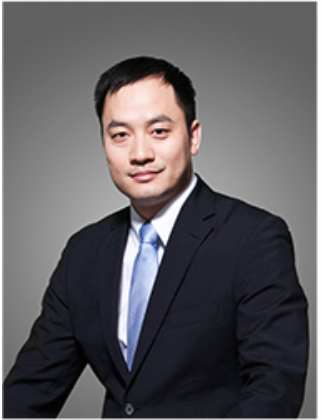}}]{Yanfeng Wang} is currently a Professor in School of Artificial Intelligence, Shanghai Jiao
Tong University. He received the B.S. degree from PLA Information Engineering University, Beijing, China, and the M.S. and Ph.D. degrees in business management from Shanghai Jiao Tong University, Shanghai, China. He is currently the Vice Director of Cooperative Medianet Innovation Center and also the Vice Dean of the School of Electrical and Information Engineering, Shanghai Jiao Tong University. His research interest mainly include media big data, the emerging commercial applications of information technology, and technology transfer.
\end{IEEEbiography}




\end{document}